\documentclass[11pt]{article}

\usepackage[preprint]{acl}

\usepackage{times}
\usepackage{latexsym}

\usepackage[T1]{fontenc}

\usepackage[utf8]{inputenc}

\usepackage{microtype}

\usepackage{inconsolata}

\usepackage{graphicx}

\usepackage{subcaption}
\usepackage{float}
\usepackage{amsmath}
\usepackage{tcolorbox}
\usepackage{mdframed}
\usepackage{caption}
\usepackage{xcolor}
\usepackage{pifont}
\usepackage{booktabs}
\usepackage{stfloats}
\usepackage{url}
\usepackage{graphicx}
\usepackage{enumitem}
\usepackage[ruled,vlined]{algorithm2e}
\usepackage{soul}

\definecolor{syscolor}{RGB}{0,100,0}
\definecolor{btcolor}{RGB}{0,0,180}
\definecolor{bicolor}{RGB}{150,0,150}
\definecolor{hcolor}{RGB}{80,80,80}
\definecolor{ytcolor}{RGB}{180,0,0}
\definecolor{dialoguebg}{RGB}{248,248,248}
\definecolor{dialogueframe}{RGB}{180,180,180}
\definecolor{copiedcolor}{RGB}{200,0,0}

\newcommand{\copied}[1]{\begingroup\sethlcolor{yellow!35}\hl{#1}\endgroup}

\newcommand{\spkone}[1]{\textbf{Speaker 1:} \textit{#1}\\[0.3em]}
\newcommand{\spktwo}[1]{\textbf{Speaker 2:} \textit{#1}\\[0.3em]}

\newcommand{\Strangers}{\textit{Strangers}}      %
\newcommand{\Fan}{\textit{Fan}}          %
\newcommand{\Peers}{\textit{Peers}}              %

\title{Stranger, Fan, or Peer? A Systematic Study on the Role of Interlocutor in Persona-Based Dialogue Generation}

\author{
 Daniela Occhipinti$^{1}$,
 Malvina Nissim$^2$,
 Marco Guerini$^1$
 \\
 $^1$Fondazione Bruno Kessler, Via Sommarive 18, Povo, Trento, Italy\\
 $^2$University of Groningen, Netherlands
 \\
 \texttt{docchipinti@fbk.eu, m.nissim@rug.nl, guerini@fbk.eu}
}

\begin{document}
\maketitle
\begin{abstract}
Persona-based dialogue systems are usually conditioned on speaker biography, but dialogues involve at least two participants, and \textit{who} has access to \textit{whose} biography can vary across training, inference, and evaluation. Prior work often neglected these aspects, obscuring mechanisms that only appear when biography visibility is toggled separately across training, inference, and evaluation, a three-stage factorisation that prior work has largely treated as a single factor. We study this factorisation on a dataset of dialogues paired with speakers' biographies, varying whether the target and interlocutor speakers see each other's biographies during training and inference, and using an LLM as a judge to perform author identification. 
We find that (i) training-time visibility, more than inference-time visibility, determines whether models express persona traits through dialogue or fall back on copying biographical text (a known problem/phenomenon in persona-based generation); (ii) models trained with interlocutor-biography visibility copy less target-biographical text than models trained without it, while changing visibility only at inference time has a less consistent effect; and (iii) under asymmetric disclosure, where only the interlocutor sees the target biography, target content leaks into interlocutor turns more often, and dialogues containing such traces are easier for the judge to identify, especially when interlocutor turns are visible. In the setting we study, these results suggest that biography leakage into generated turns is largely an artefact of how interlocutor visibility is configured across training and inference, and separating the three stages is necessary to observe it.

\end{abstract}

\section{Introduction}
\label{sec:intro}

\begin{figure}[ht!]
    \centering
    \includegraphics[width=1\columnwidth]{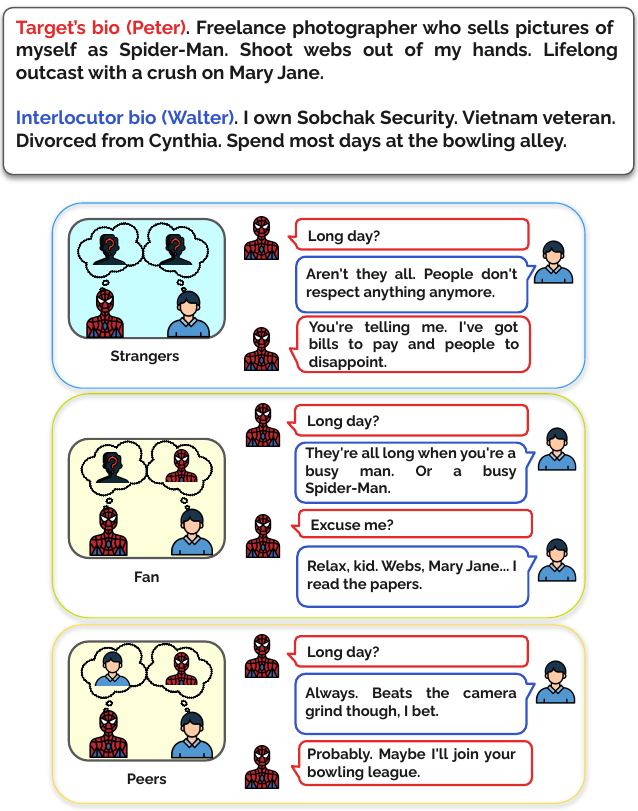}
    \caption{Three dialogues under the same biographies, varying what each model sees. (a)~Inf$_\Strangers{}$: neither model sees the other's biography.  (b)~Inf$_\Fan{}$: only the interlocutor sees the target's biography. (c)~Inf$_\Peers{}$: mutual biography disclosure.}
    \label{fig:biography-visibility}
\end{figure}

Persona-based dialogue systems condition language models on speaker biographies in order to make generated utterances more consistent and recognisable \citep{zhang-etal-2018-personalizing, liu-etal-2020-impress}. 
A key challenge that has received little attention so far is ensuring that the model meaningfully adapts to interlocutor characteristics while preserving a stable representation of the target persona, rather than degenerating into superficial copying behaviour from biographical text. Persona conditioning is not only a question of what the target speaker knows about itself: %
the interlocutor can also be conditioned on a biography, and what each agent sees about the other can differ across training, inference, and evaluation. %
This raises an under-explored question: %
\textit{at what stage of the pipeline, and for whom, does interlocutor biography visibility shape the resulting dialogue?}

Whether a speaker is identifiable from a generated dialogue depends on what information was available, and to whom, when the dialogue was produced. A speaker may be identifiable because it expresses a stable persona, because it adapts naturally to the interlocutor, or because it copies surface details from a biography. Recent work measures this adaptation through an author identification framework \citep{occhipinti-etal-2025-harry}: an LLM judge is shown a dialogue and three candidate biographies and must select the one corresponding to the target speaker. Masking or revealing interlocutor information to the judge changes identification accuracy, suggesting that interlocutor information is reflected in the generated dialogue and impacts target recognisability. However, \citet{occhipinti-etal-2025-harry} vary interlocutor visibility only at evaluation time, holding training and inference fixed. This leaves open whether exposure during training changes how the model uses interlocutor information, whether inference time disclosure has a separable effect, and whether asymmetric disclosure between participants creates biography leakage.

\begin{figure*}[t!]
    \centering
    \includegraphics[width=1\textwidth]{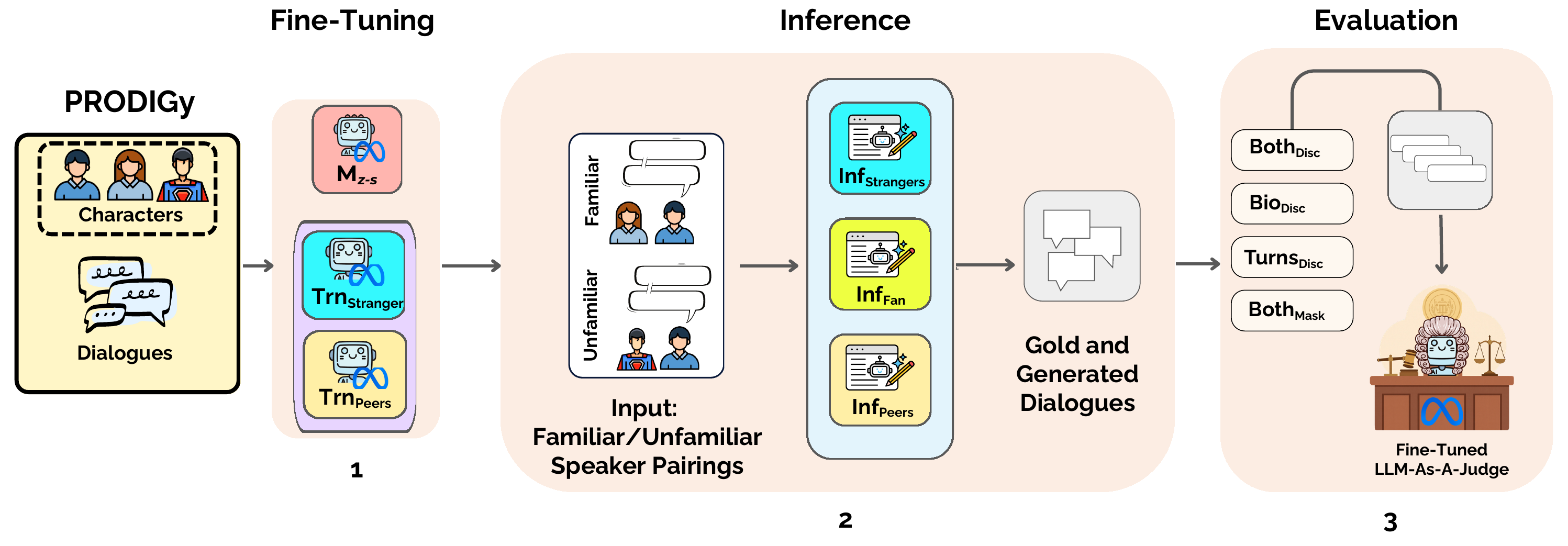}
    \caption{Pipeline. (i) Fine-tune LLaMA 3.1 8B Instruct on PRODIGy dialogues under two training-time biography visibility regimes, Trn$_\Strangers{}$ (neither speaker sees the other) and Trn$_\Peers{}$ (both speakers see each other), and keep the zero-shot backbone M$_{z\text{-}s}$ as a third regime. (ii) Generate dialogues for familiar and unfamiliar speaker pairings (see Section~\ref{sec:familiarity}) under three inference-time visibility conditions: Inf$_\Strangers{}$ (no disclosure), Inf$_\Fan{}$ (only the interlocutor sees the target), and Inf$_\Peers{}$ (mutual disclosure). (iii) Evaluate dialogues with a fine-tuned judge that performs author identification, systematically varying the interlocutor information available to the judge.}
    \label{fig:pipeline-overview}
\end{figure*}

We disentangle these stages, factorising \textit{who knows whose biography} across train, inference and evaluation.
At each stage, this is set by two binary decisions: whether the target has access to the interlocutor's biography, and whether the interlocutor has access to the target's. We study three combinations: neither speaker knows the other, both know each other, and the asymmetric case where only the interlocutor knows the target.\footnote{We omit the inverse asymmetric case because our interest concerns how target information leaks into interlocutor turns, a directional phenomenon that can affect target recognisability.} These are applied at training and inference time, while evaluation visibility (what the judge sees) is varied separately to diagnose what drives identification. Figure~\ref{fig:biography-visibility} shows the three configurations, Figure~\ref{fig:pipeline-overview} the pipeline.

This setup yields three research questions:

\begin{description}[leftmargin=0pt, labelwidth=1.5cm, labelsep=0.5em]
\itemsep0em

\item[RQ1] \textbf{Training-time and inference-time visibility.} How much does it matter, for the model's outputs, whether the target sees the interlocutor's biography during training, and whether the interlocutor's biography is supplied at inference time?

\item[RQ2] \textbf{Training--inference interaction.} When training-time and inference-time visibility are varied independently, does aligning them (concordant) produce a different behaviour compared to leaving them mismatched (discordant)?

\item[RQ3] \textbf{Target information through interlocutor turns.} When only the interlocutor knows the target's biography, do target biographical details surface in the interlocutor's turns, and do those traces make the target easier to identify?
\end{description}

Our contributions are threefold. (i) We present the first systematic study of interlocutor biography visibility as a three-stage factor, separating training, inference, and evaluation, and show that biographical copying, a known failure mode of persona-based generation, is primarily an artefact of training-time methodology rather than a property of persona conditioning. (ii) We show that training-time and inference-time visibility do not act independently: the effect of supplying the interlocutor biography at inference depends on whether the model was trained to use it. (iii) We identify biography leakage under asymmetric disclosure, in which target biographical traces surface in interlocutor turns and make target identification easier, an effect that grows when the two speakers come from unrelated narrative worlds.

\section{Related Work}
\label{sec:related}

\paragraph{Persona-based dialogue and the interlocutor.}

Persona-based dialogue systems condition models on the target speaker's biography \citep{li-etal-2016-persona, zhang-etal-2018-personalizing}, and a separate line of work conditions on the interlocutor's persona alongside the target's \citep{liu-etal-2020-impress, gu2021partner, xu2022cosplay, lu-etal-2022-partner}. These works show that partner-aware generation is helpful but do not measure how interlocutor information actually shapes the output. \citet{occhipinti-etal-2025-harry} measure this through an author identification framework: an LLM judge attributes generated dialogues to candidate target biographies, and show that masking the interlocutor's biography from the judge degrades accuracy, suggesting an impact of this information on generation. Their study, however, varies interlocutor visibility only at evaluation time.

\paragraph{Information leakage in persona-based dialogue.}

Large language models can leak information about their training data \citep{carlini2021extracting, carlini2022quantifying} and about content placed in their context window \citep{perez2022ignore}. In persona-based dialogue, this manifests as a tendency to copy biographical text into generated turns rather than expressing persona through dialogue \citep{zhang-etal-2018-personalizing, Dinan2019TheSC}. Zero-shot LLMs are especially prone to verbatim copying, which fine-tuning substantially reduces \citep{occhipinti-etal-2025-harry}. Existing work focuses on the within-speaker case, where a speaker's own biography surfaces in their own turns; the multi-agent case, in which one speaker's biography could surface in the other speaker's turns, has received less attention.

\paragraph{LLMs as evaluators and persona identifiers.}

LLMs have emerged as effective automated judges for dialogue and correlate well with human assessments \citep{chen2024mllmasajudge, gu2025surveyllmasajudge}. For authorship attribution and persona identification specifically, humans perform poorly \citep{bao-etal-2025-whos}, while LLMs outperform most humans on related tasks such as personality prediction \citep{schoenegger2025ai}. We therefore use an LLM judge throughout this work.
\section{Experimental Design}
\label{sec:methodology}

Our setup pairs two LLM agents in dialogue, each conditioned on its own biography, and varies independently across training, inference, and evaluation what each agent (and the judge) sees of the other's biography. We adopt the PRODIGy dataset and setup described in \citet{occhipinti-etal-2025-harry}, with the three-stage visibility factorisation as our extension.

\subsection{Dataset and Dialogue Setting}
\label{sec:dataset}
The PRODIGy dataset \citep{occhipinti-etal-2024-prodigy} is a movie-dialogue corpus in which characters are annotated with multiple profile dimensions (biography, communication style, personality, gender). We focus on biographies, which are short first-person narratives (see Table~\ref{tab:app-exp-RQ3-biographies} in Appendix~\ref{app:dialogue-examples} for examples), because they carry the richest semantic content for persona conditioning.

\paragraph{Data and splits.} We use the subset of 5,660 dialogues in which both speakers have biographical annotations, with the original 80:10:10 train/validation/test split, biographies truncated to the first five sentences, and the topic labels (e.g., \textit{Emotions}, \textit{Fiction}) used to condition generation.

\paragraph{Dialogue generation.} Dialogues are produced by two LLM agents, one per speaker: a target (Speaker~1, biography $B_t$) and an interlocutor (Speaker~2, biography $B_i$). Each agent receives a system prompt with its own biography and, depending on the visibility configuration (Section~\ref{subsec:visibility}), the other speaker's biography. Agents alternate for 8 turns (4 each). Training subset size is held fixed across configurations.

\subsection{Familiar vs.\ Unfamiliar Pairings}
\label{sec:familiarity}

The speaker pool extends PRODIGy with newly generated \textit{non-PRODIGy} characters to control for data contamination from publicly available movie scripts (see Appendix~\ref{app:profile-generation}). %

Following \citet{occhipinti-etal-2025-harry}, we call a pairing \textbf{unfamiliar} when the interlocutor "violates" the target's established narrative context (e.g., Walter Sobchak from \textit{The Big Lebowski} paired with Peter Parker from \textit{Spider-Man}), and \textbf{familiar} when it does not: either both characters come from the same narrative background (e.g., Peter Parker and Mary Jane Watson), or both are non-PRODIGy characters, which have no established narrative world to violate. The familiar condition therefore mixes shared narrative with shared generation source, and we interpret it accordingly.\footnote{The test set contains 1,875 familiar and 2,500 unfamiliar pairings per configuration.}

\subsection{Models and Training Objectives}
\label{sec:training-configs}

All generators and the LLM judge are based on LLaMA~3.1 8B Instruct \citep{dubey2024llama}. We train separate LoRA adapters on the frozen backbone, one per training-time biography visibility configuration. The zero-shot baseline (M$_{z\text{-}s}$) uses the backbone directly. Each fine-tuning example contains the system instruction, the target biography, the interlocutor biography (if visible under the considered configuration), the preceding turns of the dialogue history, and the final target-speaker turn (see Appendix \ref{app:hyperparams} for details). All fine-tuned models use a \textbf{masked-loss} objective: the model attends to the full prompt, but the loss is computed only on the final target turn, which is the sequence the model has to predict.
We compare this objective against standard next-token prediction in Appendix~\ref{app:loss-masking}.

\subsection{Interlocutor Biography Visibility}
\label{subsec:visibility}

We examine three biography visibility configurations, illustrated in Figure~\ref{fig:biography-visibility}:
\begin{itemize}[leftmargin=*]
    \item \textbf{\Strangers{} (I-I-)}: speakers do not know each other, no one sees the other's biography.
    \item \textbf{\Fan{} (I-I+)}: asymmetric disclosure; the interlocutor knows the target's biography, not vice versa.
    \item \textbf{\Peers{} (I+I+)}: mutual disclosure; both speakers see each other’s biography.
\end{itemize}

In the I$\pm$I$\pm$ shorthand, the first slot is the target's access to the interlocutor's biography and the second is the interlocutor's access to the target's; $+$ denotes access and $-$ its absence.

Training and inference visibility are varied separately, but not over identical sets. At training time, we instantiate the two symmetric regimes, Trn$_\Strangers{}$ and Trn$_\Peers{}$, to compare models trained without vs.\ with mutual interlocutor-biography access. The two asymmetric training regimes were also run and are reported in Appendix~\ref{app:asym-training}. At inference time, we evaluate Inf$_\Strangers{}$ and Inf$_\Peers{}$, and additionally Inf$_\Fan{}$, where only the interlocutor knows the target biography. We omit the inverse asymmetric case (target sees interlocutor but not vice versa) because our analysis of asymmetric disclosure (RQ3) concerns how target information propagates into interlocutor turns, a directional phenomenon that requires the interlocutor to be the side with biography access. The reverse asymmetric condition is provided in Appendix~\ref{app:full-results}.

Each research question uses a different slice of these configurations. RQ1 compares Trn$_\Strangers{}$ with Trn$_\Peers{}$, and separately Inf$_\Strangers{}$ with Inf$_\Peers{}$, isolating one stage at a time. RQ2 crosses the two stages, contrasting concordant training--inference pairs with discordant ones. RQ3 keeps training without interlocutor-biography access and compares Inf$_\Strangers{}$ with Inf$_\Fan{}$.

\subsection{Evaluation Framework}
\label{sec:evaluation-framework}

For evaluation, we adopt an \textbf{author identification} framework: an LLM judge receives the target speaker's turns and three candidate biographies, and must identify which biography corresponds to the target speaker. This framework is based on the idea that, if a dialogue expresses the intended speaker, the corresponding biography should be recoverable from plausible alternatives. Evaluation conditions vary what interlocutor-side information is additionally available to the judge regarding both biography and turns, for a total of four evaluation conditions. In Both$_{\text{Disc}}$, both the interlocutor biography and interlocutor turns are visible; in Bio$_{\text{Disc}}$, only the interlocutor biography is visible; in Turns$_{\text{Disc}}$, only the interlocutor turns are visible; and in Both$_{\text{Mask}}$, both interlocutor biography and interlocutor turns are masked. Distractor biographies are selected by SBERT similarity, and evaluation uses greedy decoding \cite{song2024good}.

We use an LLM judge since persona-based authorship identification has proven a hard task for humans, with poor results compared to automatic evaluation \citep{bao-etal-2025-whos, occhipinti-etal-2025-harry}; evidence also comes from other personality-related prediction tasks \citep{schoenegger2025ai}. The judge uses the same LLaMA~3.1 8B Instruct backbone as the generators, but is separately fine-tuned for the authorship identification task. We opted for a fine-tuned judge to specialise the evaluator for the domain of movie dialogues, which exhibit distinctive stylistic and structural features that our dialogue models consistently emulate. The judge is fine-tuned on gold PRODIGy instances following the same partitioning used for the generator experiments, but with instances reformatted as the authorship-identification task described above. Generated dialogues are used only for evaluation; the full judge prompt is reported in Appendix~\ref{app:judge-llm-prompts}. Masked loss is applied so that only the answer token contributes to the objective, and each training instance is presented three times with independently permuted answer options to mitigate positional bias \citep{pezeshkpour-hruschka-2024-large, zheng2024large}. 

We report judge identification \textbf{accuracy}, alongside \textbf{rare-word overlap}, i.e. the percentage of dialogues in which a biography and a turn set share at least one rare word. Throughout this work, \emph{rare-word overlap} is what we count and \emph{copying} is the behaviour we infer from it. The two are not the same: Section~\ref{sec:RQ3-results} reports cases where identification improves with no rare-word overlap at all.
Rarity is defined via \texttt{wordfreq}'s \texttt{zipf\_frequency} \citep{van2014subtlex}\footnote{\url{https://github.com/rspeer/wordfreq/}}, a base-10 log scale of occurrences per billion words, with threshold Zipf~$<$~4.0 (fewer than ten occurrences per million words).
We measure overlap in three directions:
\begin{itemize}[leftmargin=*]
\itemsep0em
    \item Bio$_\text{Trg}$$\rightarrow$Turn$_\text{Trg}$: target biography content in target turns.
    \item Bio$_\text{Int}$$\rightarrow$Turn$_\text{Trg}$: interlocutor biography content in target turns.
    \item Bio$_\text{Trg}$$\rightarrow$Turn$_\text{Int}$: target biography content in interlocutor turns.
\end{itemize}

\section{Results}
\label{sec:results}

We report the LLM judge's performance on gold PRODIGy dialogues (Section~\ref{sec:gold-dialogues}) as a quality reference, then address the three research questions in turn: training-time and inference-time visibility (RQ1, Section~\ref{sec:RQ1-results}), training--inference interaction (RQ2, Section~\ref{sec:RQ2-results}), and target information leakage through interlocutor turns under asymmetric disclosure (RQ3, Section~\ref{sec:RQ3-results}). Comprehensive per-configuration results are reported in Appendix~\ref{app:full-results}. 

\subsection{Gold Dialogues}
\label{sec:gold-dialogues}

\paragraph{Judge accuracy on human-written dialogues and interlocutor role.}

We give the LLM judge PRODIGy gold dialogues under the four evaluation conditions described above. This allows us to test whether the judge can recover the target speaker's biography from human dialogues, and how much its decision depends on which aspects it sees of the interlocutor.
Table~\ref{tab:gold-merged-side} reports judge accuracy and rare-word overlap on the gold PRODIGy dialogues. Accuracy is highest with full interlocutor information disclosure (Both$_\text{Disc}$: 0.914), remains high when only the biography is disclosed (Bio$_\text{Disc}$: 0.885), and drops when the biography is masked (Turns$_\text{Disc}$: 0.667; Both$_\text{Mask}$: 0.621). %
This provides evidence that the judge can recover target biographies in this setting: on human-written dialogues, where the correct answer is known and no model-generated text is involved, it identifies the target well above chance (0.333) and does so with biography--turn overlap as low as gold text allows. It also shows that interlocutor-side information contributes to identification even in human dialogue. Rare-word overlap is low across all directions, as expected for human text, and serves as our reference target for generated dialogues.

\begin{table}[ht!]
\centering
\small
\begin{tabular}{@{}lc@{}}
\toprule
\multicolumn{2}{c}{\textit{(a) Judge accuracy by evaluation condition}} \\
\midrule
Both$_\text{Disc}$  & \textbf{0.914} \\
Bio$_\text{Disc}$   & 0.885 \\
Turns$_\text{Disc}$ & 0.667 \\
Both$_\text{Mask}$  & 0.621 \\
\midrule
\multicolumn{2}{c}{\textit{(b) Rare-word overlap (\%) by direction}} \\
\midrule
Bio$_\text{Trg}$$\rightarrow$Turn$_\text{Trg}$ & 5.84 \\
Bio$_\text{Int}$$\rightarrow$Turn$_\text{Trg}$ & 5.76 \\
Bio$_\text{Trg}$$\rightarrow$Turn$_\text{Int}$ & 2.56 \\
\bottomrule
\end{tabular}
\caption{Judge accuracy on gold PRODIGy dialogues and rare-word overlaps.}
\label{tab:gold-merged-side}
\end{table}

\subsection{RQ1: Training-time and Inference-time Visibility}
\label{sec:RQ1-results}

Table~\ref{tab:RQ1-master} reports the full training $\times$ inference $\times$ evaluation grid for the \Strangers{} and \Peers{} configurations; the \Fan{} configuration is covered in Section~\ref{sec:RQ3-results}. Full tables are in Appendix~\ref{app:full-results}.

\begin{table*}[t]
\centering
\small
\begin{tabular}{@{}l rr rr rr@{}}
\toprule
& \multicolumn{2}{c}{\textbf{M$_{z-s}$}} & \multicolumn{2}{c}{\textbf{Trn$_\Strangers{}$}} & \multicolumn{2}{c}{\textbf{Trn$_\Peers{}$}} \\
\cmidrule(lr){2-3} \cmidrule(lr){4-5} \cmidrule(lr){6-7}
& \textbf{Inf$_\Strangers{}$} & \textbf{Inf$_\Peers{}$} & \textbf{Inf$_\Strangers{}$} & \textbf{Inf$_\Peers{}$} & \textbf{Inf$_\Strangers{}$} & \textbf{Inf$_\Peers{}$} \\
\midrule
\multicolumn{7}{l}{\textit{Accuracy}} \\
\quad Both$_\text{Disc}$                       & 0.936          & \textbf{0.953} & 0.830 & 0.854 & 0.803 & 0.831 \\
\quad Bio$_\text{Disc}$                        & \textbf{0.963} & 0.957          & 0.848 & 0.834 & 0.814 & 0.793 \\
\quad Turns$_\text{Disc}$                      & 0.939          & \textbf{0.947} & 0.789 & 0.818 & 0.746 & 0.782 \\
\quad Both$_\text{Mask}$                       & \textbf{0.971} & 0.962          & 0.760 & 0.765 & 0.683 & 0.672 \\
\midrule
\multicolumn{7}{l}{\textit{\% Rare-Word Overlap}} \\
\quad Bio$_\text{Trg}$$\rightarrow$Turn$_\text{Trg}$ & 41.26 & 42.29 & 3.82 & 4.00 & 2.95 & \textbf{2.65} \\\quad Bio$_\text{Int}$$\rightarrow$Turn$_\text{Trg}$ & 27.84 & 31.70 & 1.46 & 2.06 & \textbf{1.35} & 1.69 \\
\quad Bio$_\text{Trg}$$\rightarrow$Turn$_\text{Int}$ & 28.62 & 29.40 & 2.17 & 3.29 & 2.354 & \textbf{1.90} \\
\bottomrule
\end{tabular}
\caption{Identification accuracy and \% rare-word overlap across all training and inference visibility configurations and four evaluation conditions.}
\label{tab:RQ1-master}
\end{table*}

\paragraph{Training with interlocutor visibility reduces copying; inference-time visibility does not.} 

Zero-shot models reproduce target-biography words verbatim, while fine-tuning reduces this behaviour; within fine-tuned models, the remaining copying is shaped more by training-time interlocutor exposure than by inference-time disclosure.

We compare zero-shot prompting against fine-tuning, varying within fine-tuning whether the model sees the interlocutor's biography at training (Trn$_\Strangers{}$ vs.\ Trn$_\Peers{}$) and at inference (Inf$_\Strangers{}$ vs.\ Inf$_\Peers{}$). The largest contrast in biography recognition is between zero-shot prompting and fine-tuning. M$_{z-s}$ achieves the highest accuracy (0.936--0.971) but copies heavily from the target biography (Bio$_\text{Trg}$$\rightarrow$Turn$_\text{Trg}$ overlap of 41--42\%). Without fine-tuning, reproducing biography text becomes the simplest way for the model to satisfy the persona constraint. By contrast, fine-tuning brings this overlap below 5\%, in line with gold dialogues, while preserving substantial recognisability (0.672--0.854).
Within the fine-tuned configurations, training-time visibility controls residual copying. Trn$_\Peers{}$ lowers copying compared to Trn$_\Strangers{}$ under both inference settings (3.82\%$\rightarrow$2.95\% under Inf$_\Strangers{}$; 4.00\%$\rightarrow$2.65\% under Inf$_\Peers{}$), at an accuracy cost of 2--4 points (e.g., Both$_\text{Disc}$/Inf$_\Strangers{}$: 0.830 vs.\ 0.803; Both$_\text{Disc}$/Inf$_\Peers{}$: 0.854 vs.\ 0.831; Turns$_\text{Disc}$/Inf$_\Peers{}$: 0.818 vs.\ 0.782). Changing inference-time visibility with training fixed shifts accuracy by a similar amount but affects copying inconsistently: it rises slightly under Trn$_\Strangers{}$ (3.82\%$\rightarrow$4.00\%) and falls under Trn$_\Peers{}$ (2.95\%$\rightarrow$2.65\%). Thus training-time exposure to interlocutor biographies changes how the model uses this information, whereas the effect of inference-time disclosure depends on what the model learned during training.

\paragraph{Training-time visibility matters more for unfamiliar pairings.}
When the two speakers share no common background, training-time interlocutor visibility has a larger effect. The gap between Trn$_\Strangers{}$ and Trn$_\Peers{}$ widens on unfamiliar pairings for both accuracy and copying.

Under Inf$_\Strangers{}$, the accuracy advantage of Trn$_\Strangers{}$ over Trn$_\Peers{}$ is larger on unfamiliar pairings (5.7 points, 0.764 vs.\ 0.707) than on familiar ones (3.0 points, 0.865 vs.\ 0.835), and Bio$_\text{Trg}$$\rightarrow$Turn$_\text{Trg}$ overlap shows the same pattern (3.32\%$\rightarrow$4.28\% on unfamiliar pairings, 2.45\%$\rightarrow$3.20\% on familiar ones). %
Familiar pairings give the judge a second route to the target that does
not require the model to use interlocutor information: shared narrative background for same-source pairs, and a common generation source for non-PRODIGy pairs. Since unfamiliar pairings offer neither, training-time interlocutor visibility has a larger effect there on both accuracy and copying. 

\begin{table}[t]
\centering
\small
\begin{tabular}{@{}l rr rr@{}}
\toprule
& \multicolumn{2}{c}{\textbf{Trn$_\Strangers{}$}} & \multicolumn{2}{c}{\textbf{Trn$_\Peers{}$}} \\
\cmidrule(lr){2-3} \cmidrule(lr){4-5}
& \textbf{Fam.} & \textbf{Unfam.} & \textbf{Fam.} & \textbf{Unfam.} \\
\midrule
Acc (avg. evals)                               & 0.87 & 0.76 & 0.84 & 0.71 \\
Bio$_\text{Trg}$$\rightarrow$Turn$_\text{Trg}$ & 3.20 & 4.28 & \textbf{2.45} & 3.32 \\
\bottomrule
\end{tabular}
\caption{Trn$_\Strangers{}$ vs.\ Trn$_\Peers{}$ under Inf$_\Strangers{}$, by speaker familiarity. Accuracy is averaged over the four evaluation-disclosure conditions. }
\label{tab:RQ1-familiarity}
\end{table}

\paragraph{Reduced copying does not reduce dialogue quality.} Lower overlap with speakers' biographies could simply mean blander dialogue. Table~\ref{tab:quality} therefore reports four metrics that use neither the judge nor the rare-word measure: distinct-1/2 (unique unigrams and bigrams), self-BLEU (repetition within the target's turns), turn coherence (SBERT similarity between consecutive turns), and persona contradiction (the RoBERTa-large-MNLI probability \cite{liu2019roberta} of the target turns contradicting the target biography). The lower-copying Trn$_\Peers{}$ regime is \emph{more} lexically diverse, not less, closing $42\%$ of the distinct-1 gap to gold, and contradicts the biography no more often; repetition and coherence shift only slightly. Reduced copying changes what the model reuses, not how much it says.

\begin{table}[ht!]
\centering\small
\begin{tabular}{@{}l rrr@{}}
\toprule
\textbf{Metric} & \textbf{Trn$_\Strangers{}$} & \textbf{Trn$_\Peers{}$} & \textbf{Gold} \\
\midrule
distinct-1 $\uparrow$   & 0.801 & 0.833 & 0.878 \\
distinct-2 $\uparrow$   & 0.986 & 0.989 & 0.954 \\
self-BLEU $\downarrow$  & 3.89  & 4.16  & 7.15  \\
\midrule
turn coherence $\uparrow$          & 0.207 & 0.194 & 0.238 \\
persona contradiction $\downarrow$ & 0.279 & 0.278 & 0.314 \\
\bottomrule
\end{tabular}
\caption{Judge-free quality metrics at Inf$_\Strangers{}$
($n=4{,}375$ generated, $625$ gold).}
\label{tab:quality}
\end{table}

\subsection{RQ2: Training--Inference Interaction}
\label{sec:RQ2-results}

\paragraph{The effect of inference-time visibility depends on training: matched training and inference yields the lowest copying.}

When mutual disclosure is present at both stages, the model uses interlocutor information for contextual adaptation rather than target-biography copying, yielding the lowest overlap in Table~\ref{tab:RQ1-master}. 
Adding the interlocutor biography at inference raises accuracy under both training regimes (Trn$_\Strangers{}$: 0.830$\rightarrow$0.854; Trn$_\Peers{}$: 0.803$\rightarrow$0.831), but copying responds differently. Under Trn$_\Strangers{}$ (discordant) it creeps up (3.82\%$\rightarrow$4.00\%), as the model gains a recognisability cue without dropping the copying shortcut; under Trn$_\Peers{}$ (concordant) it drops (2.95\%$\rightarrow$2.65\%), as the model adapts to the interlocutor rather than reproducing the biography. Inference-time disclosure thus moves accuracy while copying behaviour is largely determined by training, so concordant Trn$_\Peers{}$/Inf$_\Peers{}$ achieves the lowest copying of any fine-tuned configuration with competitive accuracy.

\begin{figure}[ht!]
    \centering
    \includegraphics[width=0.9\columnwidth]{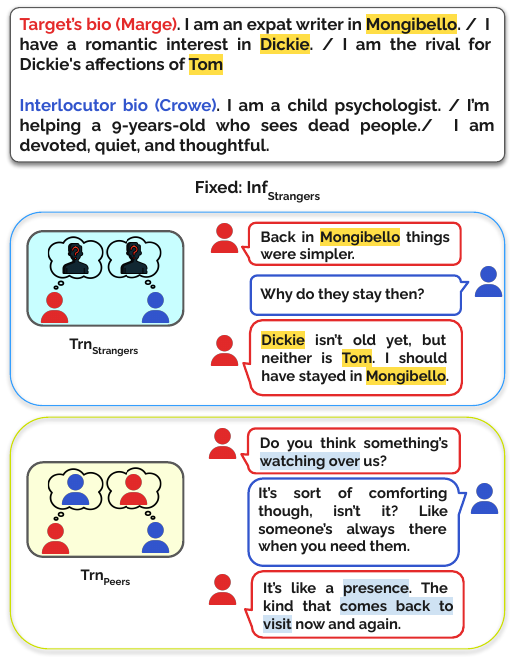}
        \caption{Marge (target) and Crowe (interlocutor) under Trn$_\Strangers{}$ vs.\ Trn$_\Peers{}$. Yellow marks exact target-biography copying; blue marks language adapted to Crowe's psychological context. Under Trn$_\Strangers{}$, Marge's turns reproduce target-biography names; under Trn$_\Peers{}$, they shift toward Crowe's context without copying Marge's biography. }
    \label{fig:RQ2-example}
\end{figure}

Figure~\ref{fig:RQ2-example} illustrates one slice of the interaction: the Trn$_\Strangers{}$ vs.\ Trn$_\Peers{}$ contrast under fixed Inf$_\Strangers{}$, with Marge Sherwood as target and Dr.\ Malcolm Crowe as interlocutor. Under Trn$_\Strangers{}$, Marge's turns reproduce target-biography terms such as \textit{Mongibello}, \textit{Dickie}, and \textit{Tom}. Under Trn$_\Peers{}$, the dialogue instead shifts toward Crowe's psychological/paranormal context, using language of presences, visitors, and unease rather than copying Marge's biographical names.

\subsection{RQ3: Information Leakage under Asymmetric Disclosure}
\label{sec:RQ3-results}
We finally ask whether target information can leak through the interlocutor's side of the dialogue. Two questions follow: does target content surface in the interlocutor's turns, and, when it does, does the judge use it? We compare Inf$_\Strangers{}$ with Inf$_\Fan{}$, in which the interlocutor sees the target biography but not vice versa. This setting tests whether target-biographical details surface in the interlocutor's generated turns, and whether those traces make the target easier to identify. We focus on Trn$_\Strangers{}$, where the model never saw the interlocutor's biography during training, so any increase in Bio$_\text{Trg}$$\rightarrow$Turn$_\text{Int}$ leakage under \Fan{} can be attributed to inference-time access alone. The zero-shot baseline provides a second regime with no training-time interlocutor knowledge, but one already dominated by the copying behaviour. Table~\ref{tab:RQ3-master} reports the full grid across both regimes.

\begin{table}[ht!]
\centering
\resizebox{\columnwidth}{!}{%
\begin{tabular}{@{}l rr rr@{}}
\toprule
& \multicolumn{2}{c}{\textbf{M$_{z-s}$}} & \multicolumn{2}{c}{\textbf{Trn$_\Strangers{}$}} \\
\cmidrule(lr){2-3} \cmidrule(lr){4-5}
& \textbf{Inf$_\Strangers{}$} & \textbf{Inf$_\Fan{}$} & \textbf{Inf$_\Strangers{}$} & \textbf{Inf$_\Fan{}$} \\
\midrule
\multicolumn{5}{l}{\textit{Accuracy}} \\
\quad Both$_\text{Disc}$                             & 0.936          & \textbf{0.952} & 0.830          & \textbf{0.873} \\
\quad Bio$_\text{Disc}$                              & 0.963          & \textbf{0.965} & \textbf{0.848} & 0.847          \\
\quad Turns$_\text{Disc}$                            & 0.939          & \textbf{0.954} & 0.789          & \textbf{0.843} \\
\quad Both$_\text{Mask}$                             & \textbf{0.971} & \textbf{0.971} & 0.760          & \textbf{0.779} \\
\midrule
\multicolumn{5}{l}{\textit{\% Rare-Word Overlap}} \\
\quad Bio$_\text{Trg}$$\rightarrow$Turn$_\text{Trg}$ & 41.26          & 42.19          & \textbf{3.82}           & 4.57           \\
\quad Bio$_\text{Int}$$\rightarrow$Turn$_\text{Trg}$ & 27.84          & 24.32          & \textbf{1.46}           & 1.62           \\
\quad Bio$_\text{Trg}$$\rightarrow$Turn$_\text{Int}$ & 28.62          & 31.61          & \textbf{2.17}           & 3.09           \\
\bottomrule
\end{tabular}}
\caption{Identification accuracy and \% rare-word overlap under \Strangers{} (Inf$_\Strangers{}$) vs.\ \Fan{} (Inf$_\Fan{}$) for both M$_{z-s}$ and Trn$_\Strangers{}$, across all evaluation conditions.}
\label{tab:RQ3-master}
\end{table}

\paragraph{Target details surface in interlocutor turns.}
When the interlocutor is given the target's biography, traces of it leak into the interlocutor’s turns, and the judge uses those traces when the turns are visible. Under \Fan{}, the percentage of dialogues in which the interlocutor's turns contain at least one rare word from the target biography increases for both M$_{z-s}$ and Trn$_\Strangers{}$: from 2.17\% to 3.09\% under Trn$_\Strangers{}$, and from 28.62\% to 31.61\% under M$_{z-s}$. 
Giving the interlocutor access to the target biography makes target-biographical traces more likely to appear in the interlocutor's turns. The accuracy rows in Table~\ref{tab:RQ3-master} are consistent with this interpretation. Under Trn$_\Strangers{}$, \Fan{} improves accuracy most clearly when the judge can see interlocutor turns (Both$_\text{Disc}$: 0.830$\rightarrow$0.873; Turns$_\text{Disc}$: 0.789$\rightarrow$0.843), while Bio$_\text{Disc}$ remains essentially unchanged (0.848$\rightarrow$0.847). 
The interlocutor's biography by itself does not give the judge a clearer view of the target; the turns do. For M$_{z-s}$, accuracy is already close to ceiling, so the additional effect of \Fan{} is smaller.

\begin{figure}[ht!]
    \centering
    \includegraphics[width=0.9\columnwidth]{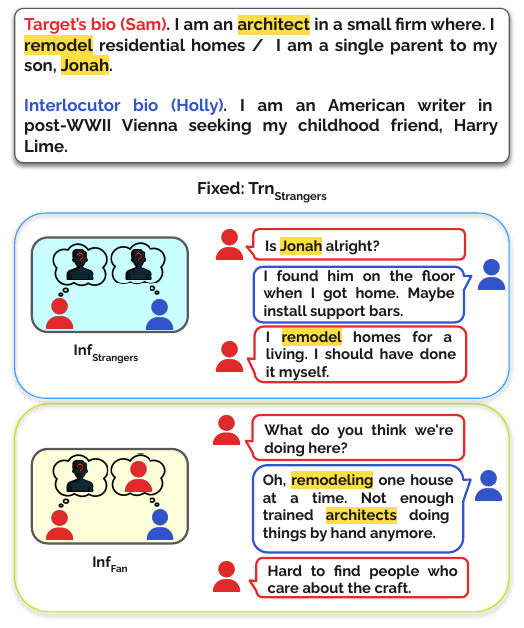}
    \caption{Sam (target) and Holly (interlocutor) under \Strangers{} vs.\ \Fan{} disclosure at inference time, with training fixed at Trn$_{\Strangers{}}$. Yellow marks rare words copied verbatim from Sam's biography. Under \Strangers{}, these appear only in Sam's turns; under \Fan{}, target-biography terms surface in Holly's turns as well.}
    \label{fig:RQ3-example}
\end{figure}

Figure~\ref{fig:RQ3-example} illustrates the mechanism on a single pairing: Sam from \textit{Sleepless in Seattle} as target and Holly Martins from \textit{The Third Man} as interlocutor. Sam's biography includes ``I am an architect in a small firm where I remodel residential homes.'' Under \Strangers{}, Holly's model has no access to Sam's biography, and biographical terms appear only in Sam's own turns. Under \Fan{}, Holly's model has access to Sam's biography and introduces architecture-related terms into its own turns (\textit{remodeling}, \textit{building}, \textit{architects}), which trace back to that biographical sentence. 

\paragraph{Rare-word leakage makes the target easier to identify.}

When rare words from the target's biography surface in the interlocutor's turns, the judge identifies the target more reliably. This holds in every configuration, confirming that the leakage measured above is informative for identification: when it occurs, the judge relies on it to recognise the target speaker. For each configuration, we split dialogues by whether the interlocutor’s turns contain at least one rare word from the target biography.

\begin{figure}[t]
\centering
\includegraphics[width=\columnwidth]{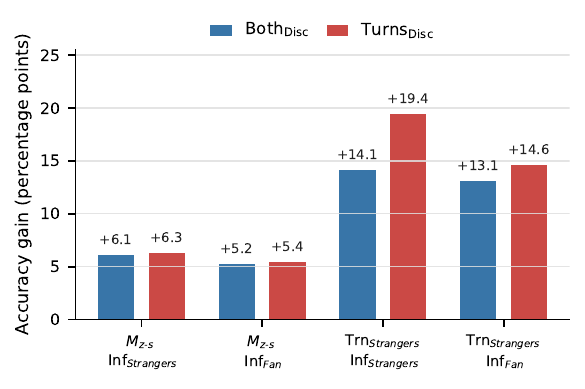}
\caption{Judge accuracy gain (percentage points) on dialogues with target biography rare-word overlap in interlocutor turns, within each configuration.}
\label{fig:RQ3-uplift}
\end{figure}

Figure~\ref{fig:RQ3-uplift} reports the accuracy gain between the two subsets. The gain is positive across all four configurations, ranging from 5.2 to 19.4 percentage points, and is larger under Turns$_\text{Disc}$ than under Both$_\text{Disc}$, where rare words in the interlocutor's turns carry greater marginal weight for the judge. The gain is also larger for fine-tuned models than for zero-shot generation, which is already near ceiling. The two trained configurations show comparable per-dialogue gains: what distinguishes \Fan{} is not the usefulness of overlap, but its higher frequency.
The no-overlap subset suggests that leakage also occurs through signals beyond exact rare-word overlap. Under Turns$_\text{Disc}$, Trn$_\Strangers{}$ accuracy on dialogues with no overlap is higher under \Fan{} than under \Strangers{} (0.839 vs.\ 0.785), suggesting asymmetric disclosure also affects interlocutor turns through paraphrase, topic steering, or more implicit semantic uptake. Full subset accuracies are reported in Appendix~\ref{app:RQ3-subsets} (Table~\ref{tab:RQ3-uplift-subsets}). 

This effect is confined to models trained without interlocutor biographies: having also trained the two asymmetric regimes, we find the \Fan{} increase in leakage is significant only under Trn$_\Strangers{}$ ($+0.91$ pp, $p=0.007$) and undetectable under the other three (all $p>0.3$). Mutual disclosure at training lowers leakage: at Inf$_\Fan{}$, Trn$_\Peers{}$ leaks $0.91$ pp less than Trn$_\Strangers{}$ ($p=0.006$). More details are provided in Appendix~\ref{app:asym-training}.

\paragraph{The main effects are statistically significant.}
With model, training regime, and evaluation condition fixed, each of the 4,375 test dialogues gives a matched \Strangers{}/\Fan{} pair, so we test correctness with two-sided exact McNemar tests and report 95\% confidence intervals from 10,000 bootstrap resamples Table~\ref{tab:significance}). \Fan{} significantly raises accuracy whenever the judge sees the interlocutor's turns, but not under Bio$_\text{Disc}$, where those turns are withheld. Rare-word overlap rises by a small but significant margin.

\begin{table}[t]
\centering\small
\setlength{\tabcolsep}{4pt}
\begin{tabular}{@{}l r c r@{}}
\toprule
\textbf{Comparison} & \textbf{Effect} & \textbf{95\% CI} & \textbf{$p$} \\
\midrule
\multicolumn{4}{@{}l}{\textit{\Fan{} effect on accuracy}} \\
\quad Both$_\text{Disc}$   & $+4.30$ pp & [3.04, 5.53]  & $1.9\times10^{-11}$ \\
\quad Turns$_\text{Disc}$  & $+5.44$ pp & [4.02, 6.86]  & $1.8\times10^{-13}$ \\
\quad Bio$_\text{Disc}$    & $-0.16$ pp & [$-$1.35, 1.03] & 0.820 \\
\midrule
\multicolumn{4}{@{}l}{\textit{\Fan{} effect on leakage}} \\
\quad Bio$_\text{Trg}$$\rightarrow$Turn$_\text{Int}$ & $+0.91$ pp & [0.27, 1.58] & 0.007 \\
\bottomrule
\end{tabular}
\caption{Paired tests for the Trn$_\Strangers{}$,
Inf$_\Strangers{}\rightarrow$Inf$_\Fan{}$ comparison ($n=4{,}375$).}
\label{tab:significance}
\end{table}

\paragraph{Leakage is amplified for unfamiliar pairings.}

When the two speakers do not share a background context, asymmetric disclosure pushes more target information into the interlocutor's turns and yields a larger accuracy gain. Familiarity dampens both the leakage increase and the accuracy gain.
Table~\ref{tab:RQ3-familiarity} breaks down rare-word leakage and average judge accuracy under Trn$_\Strangers{}$ by speaker familiarity, with accuracy averaged over the four evaluation-disclosure conditions. Moving from \Strangers{} to \Fan{} raises Bio$_\text{Trg}$$\rightarrow$Turn$_\text{Int}$ leakage by 0.6 points on familiar pairings (2.56\%$\rightarrow$3.20\%) and by 1.1 points on unfamiliar ones (1.88\%$\rightarrow$3.00\%). Average accuracy also rises more for unfamiliar pairings (0.76$\rightarrow$0.81) than for familiar ones (0.87$\rightarrow$0.88). This suggests that, when speakers lack shared narrative context, the interlocutor model relies more on the target-specific information made available at inference time, and those traces become a stronger identification cue.

\begin{table}[t]
\centering
\small
\setlength{\tabcolsep}{3.5pt}
\begin{tabular}{@{}lcccc@{}}
\toprule
& \multicolumn{2}{c}{\textbf{Familiar}} 
& \multicolumn{2}{c}{\textbf{Unfamiliar}} \\
\cmidrule(lr){2-3} \cmidrule(l){4-5}
& Inf$_\Strangers{}$ & Inf$_\Fan{}$ & Inf$_\Strangers{}$ & Inf$_\Fan{}$ \\
\midrule
Acc. (avg. eval.)      & 0.865 & \textbf{0.877} & 0.763 & 0.805 \\
Bio$_\text{Trg}$$\rightarrow$Turn$_\text{Int}$  & 2.560 & 3.200 & \textbf{1.880} & 3.000 \\
\bottomrule
\end{tabular}
\caption{Bio$_\text{Trg}$$\rightarrow$Turn$_\text{Int}$ rare-word leakage and judge accuracy under Trn$_\Strangers{}$, split by speaker familiarity. Accuracy averaged over the four evaluation conditions.}\label{tab:RQ3-familiarity}
\end{table}

\section{Conclusion}
\label{sec:conclusion}

We presented a systematic study of how varying who knows whose biography across training, inference, and evaluation affects persona-based dialogue generation.
Our main findings are threefold. First, training-time interlocutor visibility determines how models leverage that information at inference time. Models trained with mutual disclosure (Trn$_{\Peers{}}$) learn role separation and use interlocutor information for contextual adaptation; models trained without it (Trn$_{\Strangers{}}$) resort to compensatory copying when that information is introduced at inference time. When the goal is persona expression rather than biography reproduction, this favours training with interlocutor-biography visibility and masked loss, which reduces copying while preserving recognisability (Table~\ref{tab:RQ1-master}; Appendix~\ref{app:loss-masking}).
Second, training-time visibility is the stronger driver of copying behaviour. Training with interlocutor biographies reduces target-biography copying more reliably than adding or removing interlocutor biographies only at inference time. Deployment-time changes alone are therefore insufficient: withholding biographies at inference does not reliably undo copying learned during fine-tuning.
Third, \Fan{} disclosure reveals an indirect identification channel: target biographical content propagates into interlocutor turns, creating exploitable signals for speaker recognition, an effect amplified for unfamiliar speaker pairings. This is also a privacy risk: giving one agent access to another's profile can cause that profile to surface indirectly in observable dialogue, which argues for limiting cross-party biography access in multi-agent systems.
Taken together, these results indicate that, within the setting of this work, reliance on explicit biographical reproduction is largely an artefact of training and visibility configuration rather than an inevitable consequence of persona conditioning.

\section*{Limitations}

We use PRODIGy because our visibility factorisation requires
structured biographies for both speakers; without this, the
\Strangers{}/\Fan{}/\Peers{} conditions cannot be instantiated. It features
movie dialogues and fictional characters, which may introduce stereotyped roles
and inherent biases, and we agree that using a single dataset of movie-dialogues 
limits generality. The mechanism we study, how biography visibility across
training, inference, and evaluation affects identifiability, is not inherently
tied to movie dialogue, but its strength may vary across domains, profile
formats, and conversational settings. We leave cross-domain validation to future
work.

We use a single backbone, LLaMA~3.1 8B Instruct, because our experimental design isolates the effects of biography-visibility settings while holding architecture and scale constant. The patterns are not tied to one
training variant, since they appear across zero-shot and fine-tuned generators and in the masked-loss ablation. However, we do not claim that the exact effect sizes will transfer unchanged to other model families or scales.

Rare-word overlap captures exact lexical transfer, not paraphrase or other semantic uptake, and is therefore a lower bound on leakage. Our evaluation uses an LLM as a judge, validated on held-out human-written dialogues against deliberately hard distractors and complemented by judge-free quality metrics (Table~\ref{tab:quality}), but it remains an automatic proxy for human
judgement.

\section*{Ethics Statement}

Role-play agents carry several risks, including toxicity, bias, hallucinations, and privacy violations \citep{chen2024persona}. Early studies have revealed a tendency of these models to generate harmful content \citep{wen-etal-2023-unveiling}, which not only degrades the user experience but also presents significant safety concerns. Furthermore, they often exhibit role-based biases, stemming from both inherent biases in their pre-training data and user prompts that may inadvertently direct them toward biased outputs \citep{perez2022ignore, branch2022evaluating}. Additionally, these systems might exhibit character hallucination \citep{ahn-etal-2024-timechara}, producing responses that do not align with their assigned roles. Lastly, role-play systems may pose privacy risks by inadvertently disclosing users' private information \citep{krishnamurthy2011privacy, corrigan2014does}. Our work on interlocutor disclosure surfaces an additional privacy-adjacent concern: under asymmetric disclosure, target biographical content can propagate into interlocutor turns, creating an indirect channel through which persona information may leak in conversational settings.

\section*{Acknowledgments}
This work was partly supported by the   ECLIPSE project - PrEventing and Combating onLine and offline hate speech and dIsinformation through multidisciPlinary innovation, education, and awareneSs in Europe (funded under the Horizon Europe Programme, grant ID: 101225823).

\bibliography{anthology}

\appendix

\section{Implementation Details}
\label{app:hyperparams}

Table~\ref{tab:app-hyperparams} lists the settings used for all
fine-tuned generators, for dialogue generation, and for the judge. All
configurations share these values; only biography visibility differs between
them.

\begin{table}[ht!]
\centering\small
\begin{tabular}{@{}ll@{}}
\toprule
\textbf{Setting} & \textbf{Value} \\
\midrule
\multicolumn{2}{@{}l}{\textit{Fine-tuning}} \\
\quad LoRA        & rank 16, alpha 32, dropout 0 \\
\quad Optimizer   & AdamW, lr 2e-5, weight decay 0.01 \\
\quad Schedule    & cosine, warmup 0.03 \\
\quad Batch       & size 4 $\times$ grad-accum 8, 3 GPUs \\
\quad Training    & up to 10 epochs (early stopping) \\
\quad Seed        & 0 \\
\midrule
\multicolumn{2}{@{}l}{\textit{Dialogue generation}} \\
\quad Decoding    & sampling, temp.\ 0.8, top-$p$ 0.9 \\
\quad Rep.\ penalty & 1.2 \\
\quad Length      & 128 tokens/turn, 8 turns \\
\midrule
\multicolumn{2}{@{}l}{\textit{Judge}} \\
\quad Decoding    & greedy \\
\bottomrule
\end{tabular}
\caption{Training, generation, and evaluation settings, held fixed
across all visibility configurations.}
\label{tab:app-hyperparams}
\end{table}

\section{Non-PRODIGy Character Generation}
\label{app:profile-generation}

To control for data contamination from publicly available movie scripts, we supplement PRODIGy with newly generated \textit{non-PRODIGy} characters. Biographies are produced with GPT-4o-mini \citep{openai2024gpt4omini} from entries in the Persona Hub dataset \citep{ge2024scaling}, a large-scale collection of role-based persona descriptions. Each Persona Hub entry is paired with a randomly assigned gender and Myers-Briggs Type Indicator (MBTI) personality type, then passed to GPT-4o-mini with the prompt in Table~\ref{tab:prompt-new-characters}. The prompt instructs the model to produce a ten-sentence first-person biography following the PRODIGy format, covering occupation, relationship status, lifestyle, and family background. Unlike the original PRODIGy biographies, which portray fictional movie characters, the resulting biographies represent ordinary individuals in everyday contexts.

\begin{table}[ht!]
\small
\begin{mdframed}[backgroundcolor=dialoguebg, linecolor=dialogueframe, linewidth=1.2pt,
  roundcorner=5pt, innerleftmargin=10pt, innerrightmargin=10pt,
  innertopmargin=8pt, innerbottommargin=8pt]
\ttfamily\footnotesize\raggedright
Considering the following persona sentence:\\[0.3em]
\mbox{}\quad \textbf{[persona sentence]}\\[0.3em]
the following gender:\\[0.3em]
\mbox{}\quad \textbf{[gender]}\\[0.3em]
and the following mbti:\\[0.3em]
\mbox{}\quad \textbf{[mbti]}\\[0.3em]
please create a profile of a person, using the following structure:\\[0.3em]
\mbox{}\quad \textbf{\{\{"gender": "gender",}\\
\mbox{}\quad \textbf{"mbti": "mbti",}\\
\mbox{}\quad \textbf{"biography": [}\\
\mbox{}\quad \textbf{"sentence 1",}\\
\mbox{}\quad \textbf{"sentence 2",}\\
\mbox{}\quad \textbf{".",}\\
\mbox{}\quad \textbf{"sentence 10"}\\
\mbox{}\quad \textbf{]\}\}}\\[0.3em]
Please ensure the biography contains up to 10 sentences in the first person singular. Include details about the individual's job, relationship status, lifestyle, and family background. Be sure to capture a varied portrayal of the individual's life and character.\\[0.3em]
Please ensure to provide the dictionary only, without anything else.
\end{mdframed}
\caption{Prompt used to generate non-PRODIGy biographies with GPT-4o-mini. Placeholders \textbf{[persona sentence]}, \textbf{[gender]}, and \textbf{[mbti]} are filled per Persona Hub entry.}
\label{tab:prompt-new-characters}
\end{table}

\section{Loss Masking: Objective and Ablation}
\label{app:loss-masking}

This section motivates the masked-loss objective (MaskLoss, ML) used in all fine-tuned models in the main paper. We compare it with standard next-token prediction (NoMaskLoss, Trn$_{\neg\text{ML}}$) and formalise the final-turn objective.
 
\subsection{Training Objectives}
 
We compare two fine-tuning objectives. NoMaskLoss applies standard next-token prediction over all tokens in the training sequence $X$:
\begin{equation}
    \mathcal{L}_{\neg \text{ML}} = - \sum_{j=1}^{|X|} \log p_\theta(x_j \mid x_{<j}),
\end{equation}
MaskLoss instead restricts the loss to the final target turn $Y_T$:
\begin{equation}
    \mathcal{L}_{\text{ML}} = - \sum_{j \in \text{idx}(Y_T)} \log p_\theta(x_j \mid x_{<j}),
\end{equation}
In MaskLoss, all tokens outside the final target turn are masked from the loss. Biographies and dialogue history therefore remain available as context, but they are not themselves prediction targets. We compare the two objectives under the \Peers{} configuration at both training and inference time, alongside the zero-shot baseline (M$_{z\text{-}s}$).

\subsection{Overall Comparison}

Table~\ref{tab:rRQ1.1-combined} compares zero-shot generation, NoMaskLoss, and MaskLoss under the \Peers{} configuration. In this ablation, overlap values are percentages of dialogues with at least one shared rare word. The zero-shot model achieves the highest identification accuracy, but it also shows much higher rare-word overlap with the biographies, indicating that its recognisability is strongly supported by surface reproduction. Among fine-tuned models, NoMaskLoss is slightly more accurate when the judge has access to the interlocutor biography, whereas MaskLoss achieves the same Turns$_\text{Disc}$ accuracy while reducing several forms of rare-word overlap. We therefore use MaskLoss in the main experiments: it preserves turn-based recognisability in this setting while reducing biographical copying.

\begin{table}[!htbp]
\centering
\small
\begin{tabular}{@{}l rrr@{}}
\toprule
& \textbf{M$_{z\text{-}s}$} & \textbf{Trn$_{\neg\text{ML}}$} & \textbf{ML} \\ 
\midrule
& \multicolumn{3}{c}{\textit{Accuracy}} \\
\cmidrule(lr){2-4}
Both$_{Disc}$   & \textbf{0.953} & 0.840 & 0.831 \\
Bio$_{Disc}$    & \textbf{0.957} & 0.819 & 0.793 \\
Turns$_{Disc}$  & \textbf{0.947} & 0.782 & 0.782 \\
Both$_{Mask}$   & \textbf{0.962} & 0.712 & 0.672 \\
\midrule
& \multicolumn{3}{c}{\textit{\% Rare Words}} \\
\cmidrule(lr){2-4}
Bio$_{Trg}$$\rightarrow$Turn$_{Trg}$  & 42.286 & 2.949 & \textbf{2.651} \\
Bio$_{Int}$$\rightarrow$Turn$_{Trg}$   & 31.703 & \textbf{1.349} & 1.691 \\
Bio$_{Trg}$$\rightarrow$Turn$_{Int}$  & 29.394 & 2.629 & \textbf{1.897} \\
\bottomrule
\end{tabular}
\caption{Loss-objective ablation under \Peers{} training and inference.}%
\label{tab:rRQ1.1-combined}
\end{table}

\subsection{Breakdown by Speaker Familiarity}
 
The speaker-familiarity breakdown shows where this trade-off is clearest. For unfamiliar speakers, NoMaskLoss attains higher accuracy (0.741 vs.\ 0.711), but this gain coincides with higher target-biography copying in the target turns (3.200\% vs.\ 2.520\%). MaskLoss therefore trades some accuracy for lower surface-form reproduction, especially when memorised or source-specific cues are least available.
 
\begin{table}[!htbp]
\centering
\small
\begin{tabular}{@{}l rrr@{}}
\toprule
& \textbf{M$_{z\text{-}s}$} & \textbf{Trn$_{\neg\text{ML}}$} & \textbf{ML} \\ 
\midrule
Familiar (Acc.)    & \textbf{0.990} & 0.852 & 0.848 \\
Unfamiliar (Acc.)  & \textbf{0.928} & 0.741 & 0.711 \\
\midrule
\multicolumn{4}{l}{\textit{\% Rare Words — Familiar}} \\
\midrule
Bio$_{Trg}$$\rightarrow$Turn$_{Trg}$  & 42.080 & \textbf{2.773} & 2.827 \\
Bio$_{Int}$$\rightarrow$Turn$_{Trg}$   & 38.347 & 2.613 & \textbf{2.400} \\
Bio$_{Trg}$$\rightarrow$Turn$_{Int}$  & 28.587 & 3.307 & \textbf{2.453} \\
\midrule
\multicolumn{4}{l}{\textit{\% Rare Words — Unfamiliar}} \\
\midrule
Bio$_{Trg}$$\rightarrow$Turn$_{Trg}$  & 42.440 & 3.200 & \textbf{2.520} \\
Bio$_{Int}$$\rightarrow$Turn$_{Trg}$   & 26.720 & \textbf{0.400} & 1.160 \\
Bio$_{Trg}$$\rightarrow$Turn$_{Int}$  & 30.000 & 2.120 & \textbf{1.480} \\
\bottomrule
\end{tabular}
\caption{Loss-objective ablation by speaker familiarity under \Peers{} training and inference.}%
\label{tab:rRQ1.1-fam-combined}
\end{table}

\section{Judge LLM Prompts}
\label{app:judge-llm-prompts}
 
In all evaluation conditions, the judge selects the target speaker's biography from three candidates. The four variants differ only in whether the interlocutor biography is supplied and whether the interlocutor turns remain visible. Table~\ref{tab:judge-visibility-conditions} gives the visibility grid, and Table~\ref{tab:judge-base-prompt} gives the shared prompt template.

\begin{table}[!htbp]
\centering
\small
\begin{tabular}{@{}lcc@{}}
\toprule
\textbf{Eval. condition} & \textbf{Interlocutor bio} & \textbf{Interlocutor turns} \\
\midrule
Both$_\text{Disc}$  & visible & visible \\
Bio$_\text{Disc}$   & visible & masked \\
Turns$_\text{Disc}$ & masked  & visible \\
Both$_\text{Mask}$  & masked  & masked \\
\bottomrule
\end{tabular}
\caption{Information visible to the judge in each evaluation condition.}
\label{tab:judge-visibility-conditions}
\end{table}

\begin{table}[ht!]
\small
\begin{mdframed}[backgroundcolor=dialoguebg, linecolor=dialogueframe, linewidth=1.2pt,
  roundcorner=5pt, innerleftmargin=10pt, innerrightmargin=10pt,
  innertopmargin=8pt, innerbottommargin=8pt]
\ttfamily\footnotesize\raggedright
[If interlocutor biography is visible:]\\
You know that interlocutor's biography is as follows:\\
\mbox{}\quad [interlocutor's biography]\\[0.5em]
Given [the interlocutor's biography and] the following dialogue about the topic [topic] [in which the interlocutor's turns are masked], your task is to guess which of the provided biographies corresponds to the target speaker.\\[0.3em]
The biographies are provided at the end of the dialogue.\\[0.3em]
Please provide your answer as ``Biography A'', ``Biography B'', or ``Biography C''.\\[0.3em]
Please make your guess even though the dialogue may sound a little weird or unnatural.\\[0.3em]
Your response must follow this JSON format: \{``Guess'': ``Biography X''\}\\[0.5em]
DIALOGUE\\[0.3em]
\mbox{}\quad [Dialogue, with interlocutor turns masked when required]\\[0.5em]
BIOGRAPHIES\\[0.3em]
\mbox{}\quad Biography A: [Biography A sentences]\\[0.3em]
\mbox{}\quad Biography B: [Biography B sentences]\\[0.3em]
\mbox{}\quad Biography C: [Biography C sentences]
\end{mdframed}
\caption{Base LLM-as-a-Judge prompt. Visibility-dependent clauses in square brackets are inserted or omitted according to Table~\ref{tab:judge-visibility-conditions}; placeholders such as [topic] and [Dialogue] are filled for each instance.}
\label{tab:judge-base-prompt}
\end{table}

\section{Asymmetric Training Regimes}
\label{app:asym-training}

The main text reports the two symmetric training regimes,
Trn$_\Strangers{}$ and Trn$_\Peers{}$, because RQ1 contrasts training without and
with mutual interlocutor-biography access. We also trained the two asymmetric
regimes: Trn$_\Fan{}$, in which only the interlocutor sees the target biography
during training, and its converse Trn$_{I+I-}$.
Table~\ref{tab:asym-training} reports the full grid and
Table~\ref{tab:asym-tests} the corresponding paired tests, computed exactly as
in Section~\ref{sec:RQ3-results}.

These regimes separate training-time from inference-time asymmetry,
which the main text cannot do. Two results follow. First, training under \Fan{}
rather than \Strangers{} does not itself increase leakage into the
interlocutor's turns ($-0.09$ pp, $p=0.85$) or identification accuracy
($+0.75$ pp, $p=0.21$): the disclosure has to occur at inference to matter.
Second, the inference-time effect is significant only for a model trained
without interlocutor biographies. Under Trn$_\Strangers{}$, moving inference
from \Strangers{} to \Fan{} raises leakage by $0.91$ pp ($p=0.007$); under
Trn$_\Fan{}$, Trn$_\Peers{}$, and Trn$_{I+I-}$ the same change is not
significant (all $p>0.3$). Any training configuration that supplies interlocutor
biographies appears to remove the channel that inference-time disclosure would
otherwise open.

\begin{table}[t]
\centering\small
\setlength{\tabcolsep}{3.5pt}
\begin{tabular}{@{}l r c r@{}}
\toprule
 & \textbf{$\Delta$ pp} & \textbf{95\% CI} & \textbf{$p$} \\
\midrule
\multicolumn{4}{@{}l}{\textit{Trn$_\Strangers{}\rightarrow$Trn$_\Fan{}$, at Inf$_\Fan{}$}} \\
\quad leakage & $-0.09$ & [$-$0.78, 0.59] & 0.845 \\
\quad accuracy & $+0.75$ & [$-$0.39, 1.92] & 0.214 \\
\midrule
\multicolumn{4}{@{}l}{\textit{Leakage, Inf$_\Strangers{}\rightarrow$Inf$_\Fan{}$}} \\
\quad under Trn$_\Strangers{}$ & $+0.91$ & [0.27, 1.58]  & 0.007 \\
\quad under Trn$_\Fan{}$       & $+0.37$ & [$-$0.32, 1.07] & 0.327 \\
\quad under Trn$_\Peers{}$     & $-0.18$ & [$-$0.78, 0.41] & 0.608 \\
\quad under Trn$_{I+I-}$       & $+0.16$ & [$-$0.46, 0.78] & 0.663 \\
\midrule
\multicolumn{4}{@{}l}{\textit{Leakage at Inf$_\Fan{}$, vs.\ Trn$_\Strangers{}$}} \\
\quad Trn$_\Fan{}$       & $-0.09$ & [$-$0.78, 0.59]   & 0.845 \\
\quad Trn$_{I+I-}$       & $-0.62$ & [$-$1.28, 0.02]   & 0.075 \\
\quad Trn$_\Peers{}$     & $-0.91$ & [$-$1.55, $-$0.30] & 0.006 \\
\bottomrule
\end{tabular}
\caption{Paired exact McNemar tests with bootstrap 95\% CIs
($n=4{,}375$). Leakage is Bio$_\text{Trg}$$\rightarrow$Turn$_\text{Int}$;
accuracy is Both$_\text{Disc}$.}
\label{tab:asym-tests}
\end{table}

\begin{table*}[t]
\centering\small
\begin{tabular}{@{}ll rrrr rrr@{}}
\toprule
& & \multicolumn{4}{c}{\textbf{Accuracy}} & \multicolumn{3}{c}{\textbf{\% Rare-word overlap}} \\
\cmidrule(lr){3-6}\cmidrule(lr){7-9}
\textbf{Training} & \textbf{Inference} & Both$_\text{Disc}$ & Bio$_\text{Disc}$ & Turns$_\text{Disc}$ & Both$_\text{Mask}$
& Bio$_\text{Trg}$$\rightarrow$Turn$_\text{Trg}$ & Bio$_\text{Int}$$\rightarrow$Turn$_\text{Trg}$ & Bio$_\text{Trg}$$\rightarrow$Turn$_\text{Int}$ \\
\midrule
Trn$_\Fan{}$
 & Inf$_\Strangers{}$ & 0.833 & 0.848 & 0.790 & 0.759 & 4.34 & 1.78 & 2.63 \\
 & Inf$_\Fan{}$       & 0.881 & 0.863 & 0.839 & 0.783 & 3.86 & 1.60 & 2.99 \\
 & Inf$_\Peers{}$     & 0.850 & 0.833 & 0.822 & 0.748 & 4.09 & 2.06 & 2.86 \\
\midrule
Trn$_{I+I-}$
 & Inf$_\Strangers{}$ & 0.808 & 0.811 & 0.730 & 0.690 & 3.13 & 1.19 & 2.31 \\
 & Inf$_\Fan{}$       & 0.860 & 0.826 & 0.804 & 0.713 & 3.18 & 1.10 & 2.47 \\
 & Inf$_\Peers{}$     & 0.832 & 0.805 & 0.765 & 0.690 & 2.22 & 1.55 & 2.24 \\
\bottomrule
\end{tabular}
\caption{Asymmetric training regimes ($n=4{,}375$ per cell).Trn$_\Fan{}$ trains only the interlocutor with the target biography;Trn$_{I+I-}$ is its converse.}
\label{tab:asym-training}
\end{table*}

\section{Comprehensive Performance Tables}
\label{app:comprehensive-tables}
\label{app:full-results}

This section reports the full numerical grid, grouped by training objective. The overlap columns report METEOR biography--turn overlap scores, where higher values indicate greater lexical similarity. The \textbf{Acc} columns report judge accuracy. The binary rare-word percentages used in the main-text leakage analyses are reported in the main text and in Appendix~\ref{app:RQ3-subsets}.

In addition to the three main inference settings, the appendix tables include Inf$_{I+I-}$, the reverse asymmetric condition in which the target sees the interlocutor biography but the interlocutor does not see the target biography. Speaker-familiarity rows use the definitions from Section~\ref{sec:dataset}. Topic-familiarity rows distinguish topics naturally associated with the target speaker (\textit{familiar}, e.g.\ Harry from \textit{When Harry Met Sally} discussing love) from topics outside the target's typical context (\textit{unfamiliar}, e.g.\ Harry discussing war). For each system family, the first table gives the full evaluation-condition grid, the second isolates speaker familiarity, and the third isolates topic familiarity. In the familiarity tables, accuracy is averaged over the four evaluation disclosure conditions, so the figures are comparable with Table~\ref{tab:RQ3-familiarity} in the main text; the METEOR columns are unaffected by the judge's view and are reported as measured.

\subsection{Gold Dialogue Baseline}
 
\begin{table*}[!htbp]\centering
\small
\begin{tabular}{@{}lcccc@{}}
\toprule
\textbf{Eval} & \textbf{Bio$_{Trg}$$\to$Turn$_{Trg}$} & \textbf{Bio$_{Int}$$\to$Turn$_{Trg}$} & \textbf{Bio$_{Trg}$$\to$Turn$_{Int}$} & \textbf{Acc} \\
\midrule
Both$_{Disc}$  & 0.033 & 0.033 & 0.032 & 0.914 \\
Bio$_{Disc}$   & 0.033 & 0.033 & 0.032 & 0.885 \\
Turns$_{Disc}$ & 0.033 & 0.033 & 0.032 & 0.667 \\
Both$_{Mask}$  & 0.033 & 0.033 & 0.032 & 0.621 \\
\bottomrule
\end{tabular}
\caption{Gold-dialogue overlap metrics reference.}%
\label{tab:overall-gold}
\end{table*}
 
Table~\ref{tab:overall-gold} establishes the reference point for generated dialogues. Gold dialogues combine high judge accuracy with uniformly low biography--turn overlap, showing that human-written persona signal is recognisable without heavy lexical reuse from the biographies. Accuracy is highest when the judge receives both interlocutor biography and turns, and it drops when interlocutor-side information is removed. This confirms that the judge is sensitive to interlocutor context even in human dialogue, while the low METEOR values indicate that this sensitivity is not simply driven by verbatim copying.

\subsection{MaskLoss: Trn\texorpdfstring{$_{\Strangers{}}$}{
Strangers}}
 
\begin{table*}[!htbp]
\centering
\small
\begin{tabular}{@{}llrrr r@{}}
\toprule
\textbf{Inf.} & \textbf{Eval} & \textbf{Bio$_{Trg}$$\to$Turn$_{Trg}$} & \textbf{Bio$_{Int}$$\to$Turn$_{Trg}$} & \textbf{Bio$_{Trg}$$\to$Turn$_{Int}$} & \textbf{Acc} \\
\midrule
Inf$_{\Strangers{}}$ & Both$_{Disc}$  & 0.038 & 0.042 & 0.046 & 0.830 \\
             & Bio$_{Disc}$   & 0.038 & 0.042 & 0.046 & 0.848 \\
             & Turns$_{Disc}$ & 0.038 & 0.042 & 0.046 & 0.789 \\
             & Both$_{Mask}$  & 0.038 & 0.042 & 0.046 & 0.760 \\
\addlinespace
Inf$_{\Fan}$ & Both$_{Disc}$  & 0.038 & 0.042 & 0.040 & 0.873 \\
             & Bio$_{Disc}$   & 0.038 & 0.042 & 0.040 & 0.847 \\
             & Turns$_{Disc}$ & 0.038 & 0.042 & 0.040 & 0.843 \\
             & Both$_{Mask}$  & 0.038 & 0.042 & 0.040 & 0.779 \\
\addlinespace
Inf$_{I+I-}$ & Both$_{Disc}$  & 0.039 & 0.036 & 0.046 & 0.824 \\
             & Bio$_{Disc}$   & 0.039 & 0.036 & 0.046 & 0.827 \\
             & Turns$_{Disc}$ & 0.039 & 0.036 & 0.046 & 0.766 \\
             & Both$_{Mask}$  & 0.039 & 0.036 & 0.046 & 0.737 \\
\addlinespace
Inf$_{\Peers{}}$ & Both$_{Disc}$  & 0.038 & 0.036 & 0.040 & 0.854 \\
             & Bio$_{Disc}$   & 0.038 & 0.036 & 0.040 & 0.834 \\
             & Turns$_{Disc}$ & 0.038 & 0.036 & 0.040 & 0.818 \\
             & Both$_{Mask}$  & 0.038 & 0.036 & 0.040 & 0.765 \\
\bottomrule
\end{tabular}
\caption{MaskLoss trained without interlocutor-biography visibility (Trn$_{\Strangers{}}$).}%
\label{tab:overall-wml-mii}
\end{table*}
 
\begin{table*}[!htbp]
\centering
\small
\begin{tabular}{@{}l rrrr@{}}
\toprule
& \textbf{Inf$_{\Strangers{}}$} & \textbf{Inf$_{\Fan}$} & \textbf{Inf$_{I+I-}$} & \textbf{Inf$_{\Peers{}}$} \\
\midrule
& \multicolumn{4}{c}{\textit{Accuracy (mean over eval. conditions)}} \\
\cmidrule(lr){2-5}
Familiar   & 0.865 & 0.877 & 0.864 & 0.879 \\
Unfamiliar & 0.763 & 0.804 & 0.731 & 0.772 \\
\midrule
& \multicolumn{4}{c}{\textit{METEOR — Familiar}} \\
\cmidrule(lr){2-5}
Bio$_{Trg}$$\to$Turn$_{Trg}$ & 0.038 & 0.037 & 0.038 & 0.038 \\
Bio$_{Int}$$\to$Turn$_{Trg}$ & 0.044 & 0.044 & 0.037 & 0.037 \\
Bio$_{Trg}$$\to$Turn$_{Int}$ & 0.047 & 0.041 & 0.047 & 0.041 \\
\midrule
& \multicolumn{4}{c}{\textit{METEOR — Unfamiliar}} \\
\cmidrule(lr){2-5}
Bio$_{Trg}$$\to$Turn$_{Trg}$ & 0.038 & 0.038 & 0.039 & 0.038 \\
Bio$_{Int}$$\to$Turn$_{Trg}$ & 0.042 & 0.042 & 0.036 & 0.036 \\
Bio$_{Trg}$$\to$Turn$_{Int}$ & 0.045 & 0.039 & 0.046 & 0.039 \\
\bottomrule
\end{tabular}
\caption{Speaker familiarity for MaskLoss / Trn$_{\Strangers{}}$.}%
\label{tab:fam-wml-mii}
\end{table*}
 
\begin{table*}[!htbp]
\centering
\small
\begin{tabular}{@{}l rrrr@{}}
\toprule
& \textbf{Inf$_{\Strangers{}}$} & \textbf{Inf$_{\Fan}$} & \textbf{Inf$_{I+I-}$} & \textbf{Inf$_{\Peers{}}$} \\
\midrule
& \multicolumn{4}{c}{\textit{Accuracy (mean over eval. conditions)}} \\
\cmidrule(lr){2-5}
Familiar   & 0.829 & 0.857 & 0.804 & 0.844 \\
Unfamiliar & 0.790 & 0.820 & 0.778 & 0.798 \\
\bottomrule
\end{tabular}
\caption{Topic familiarity for MaskLoss / Trn$_{\Strangers{}}$.}%
\label{tab:topic-wml-mii}
\end{table*}
 
Tables~\ref{tab:overall-wml-mii}--\ref{tab:topic-wml-mii} show the behaviour of a MaskLoss model trained without interlocutor-biography visibility. Overall accuracy is strongest when inference supplies additional interlocutor-side information, especially under \Fan{} (Inf$_{\Fan}$) and \Peers{} (Inf$_{\Peers{}}$), while the METEOR overlap columns remain close to the gold-dialogue range and vary little across evaluation conditions. The speaker-familiarity table shows that most of the remaining difficulty comes from unfamiliar speaker pairings: familiar pairs stay in a narrow band (0.864--0.879) regardless of the inference configuration,
whereas unfamiliar pairs range from 0.731 to 0.804 and so depend far more on what the interlocutor is given. Topic familiarity produces a smaller but consistent effect, with familiar topics generally easier than unfamiliar ones.

\subsection{MaskLoss: Trn\texorpdfstring{$_{\Peers{}}$}{[Peers]}}

\begin{table*}[!htbp]
\centering
\small
\begin{tabular}{@{}llrrr r@{}}
\toprule
\textbf{Inf.} & \textbf{Eval} & \textbf{Bio$_{Trg}$$\to$Turn$_{Trg}$} & \textbf{Bio$_{Int}$$\to$Turn$_{Trg}$} & \textbf{Bio$_{Trg}$$\to$Turn$_{Int}$} & \textbf{Acc} \\
\midrule
Inf$_{\Strangers{}}$ & Both$_{Disc}$  & 0.028 & 0.032 & 0.035 & 0.803 \\
             & Bio$_{Disc}$   & 0.028 & 0.032 & 0.035 & 0.814 \\
             & Turns$_{Disc}$ & 0.028 & 0.032 & 0.035 & 0.746 \\
             & Both$_{Mask}$  & 0.028 & 0.032 & 0.035 & 0.683 \\
\addlinespace
Inf$_{\Fan}$ & Both$_{Disc}$  & 0.028 & 0.032 & 0.028 & 0.849 \\
             & Bio$_{Disc}$   & 0.028 & 0.032 & 0.028 & 0.823 \\
             & Turns$_{Disc}$ & 0.028 & 0.032 & 0.028 & 0.795 \\
             & Both$_{Mask}$  & 0.028 & 0.032 & 0.028 & 0.707 \\
\addlinespace
Inf$_{I+I-}$ & Both$_{Disc}$  & 0.028 & 0.026 & 0.034 & 0.787 \\
             & Bio$_{Disc}$   & 0.028 & 0.026 & 0.034 & 0.789 \\
             & Turns$_{Disc}$ & 0.028 & 0.026 & 0.034 & 0.703 \\
             & Both$_{Mask}$  & 0.028 & 0.026 & 0.034 & 0.673 \\
\addlinespace
Inf$_{\Peers{}}$ & Both$_{Disc}$  & 0.028 & 0.026 & 0.028 & 0.831 \\
             & Bio$_{Disc}$   & 0.028 & 0.026 & 0.028 & 0.793 \\
             & Turns$_{Disc}$ & 0.028 & 0.026 & 0.028 & 0.782 \\
             & Both$_{Mask}$  & 0.028 & 0.026 & 0.028 & 0.672 \\
\bottomrule
\end{tabular}
\caption{MaskLoss trained with mutual biography disclosure (Trn$_{\Peers{}}$).}%
\label{tab:overall-wml-miii}
\end{table*}

\begin{table*}[!htbp]
\centering
\small
\begin{tabular}{@{}l rrrr@{}}
\toprule
& \textbf{Inf$_{\Strangers{}}$} & \textbf{Inf$_{\Fan}$} & \textbf{Inf$_{I+I-}$} & \textbf{Inf$_{\Peers{}}$} \\
\midrule
& \multicolumn{4}{c}{\textit{Accuracy (mean over eval. conditions)}} \\
\cmidrule(lr){2-5}
Familiar   & 0.834 & 0.845 & 0.830 & 0.847 \\
Unfamiliar & 0.707 & 0.755 & 0.669 & 0.711 \\
\midrule
& \multicolumn{4}{c}{\textit{METEOR — Familiar}} \\
\cmidrule(lr){2-5}
Bio$_{Trg}$$\to$Turn$_{Trg}$ & 0.028 & 0.028 & 0.028 & 0.028 \\
Bio$_{Int}$$\to$Turn$_{Trg}$ & 0.034 & 0.034 & 0.028 & 0.027 \\
Bio$_{Trg}$$\to$Turn$_{Int}$ & 0.036 & 0.029 & 0.035 & 0.028 \\
\midrule
& \multicolumn{4}{c}{\textit{METEOR — Unfamiliar}} \\
\cmidrule(lr){2-5}
Bio$_{Trg}$$\to$Turn$_{Trg}$ & 0.028 & 0.028 & 0.028 & 0.028 \\
Bio$_{Int}$$\to$Turn$_{Trg}$ & 0.031 & 0.031 & 0.026 & 0.025 \\
Bio$_{Trg}$$\to$Turn$_{Int}$ & 0.034 & 0.028 & 0.034 & 0.028 \\
\bottomrule
\end{tabular}
\caption{Speaker familiarity for MaskLoss / Trn$_{\Peers{}}$.}%
\label{tab:fam-wml-miii}
\end{table*}

\begin{table*}[!htbp]
\centering
\small
\begin{tabular}{@{}l rrrr@{}}
\toprule
& \textbf{Inf$_{\Strangers{}}$} & \textbf{Inf$_{\Fan}$} & \textbf{Inf$_{I+I-}$} & \textbf{Inf$_{\Peers{}}$} \\
\midrule
& \multicolumn{4}{c}{\textit{Accuracy (mean over eval. conditions)}} \\
\cmidrule(lr){2-5}
Familiar   & 0.783 & 0.819 & 0.758 & 0.794 \\
Unfamiliar & 0.745 & 0.775 & 0.723 & 0.752 \\
\bottomrule
\end{tabular}
\caption{Topic familiarity for MaskLoss / Trn$_{\Peers{}}$.}%
\label{tab:topic-wml-miii}
\end{table*}

Tables~\ref{tab:overall-wml-miii}--\ref{tab:topic-wml-miii} report the corresponding MaskLoss model trained with mutual biography disclosure. Compared with Trn$_{\Strangers{}}$, this model has lower METEOR overlap in all three directions, especially for target-biography overlap in target turns, which is the desired copying reduction. The cost is a small drop in accuracy, most visible when the model is evaluated without the interlocutor context it saw during training. The familiarity breakdowns preserve the same broad pattern as Trn$_{\Strangers{}}$: familiar speakers are easier than unfamiliar speakers, and topic familiarity is a weaker source of variation than speaker familiarity.

\subsection{NoMaskLoss: Trn\texorpdfstring{$_{\Strangers{}}$}{Strangers}}

\begin{table*}[!htbp]
\centering
\small
\begin{tabular}{@{}llrrr r@{}}
\toprule
\textbf{Inf.} & \textbf{Eval} & \textbf{Bio$_{Trg}$$\to$Turn$_{Trg}$} & \textbf{Bio$_{Int}$$\to$Turn$_{Trg}$} & \textbf{Bio$_{Trg}$$\to$Turn$_{Int}$} & \textbf{Acc} \\
\midrule
Inf$_{\Strangers{}}$ & Both$_{Disc}$  & 0.037 & 0.042 & 0.045 & 0.832 \\
             & Bio$_{Disc}$   & 0.037 & 0.042 & 0.045 & 0.844 \\
             & Turns$_{Disc}$ & 0.037 & 0.042 & 0.045 & 0.786 \\
             & Both$_{Mask}$  & 0.037 & 0.042 & 0.045 & 0.753 \\
\addlinespace
Inf$_{\Fan}$ & Both$_{Disc}$  & 0.037 & 0.041 & 0.038 & 0.873 \\
             & Bio$_{Disc}$   & 0.037 & 0.041 & 0.038 & 0.846 \\
             & Turns$_{Disc}$ & 0.037 & 0.041 & 0.038 & 0.834 \\
             & Both$_{Mask}$  & 0.037 & 0.041 & 0.038 & 0.769 \\
\addlinespace
Inf$_{I+I-}$ & Both$_{Disc}$  & 0.038 & 0.035 & 0.045 & 0.808 \\
             & Bio$_{Disc}$   & 0.038 & 0.035 & 0.045 & 0.824 \\
             & Turns$_{Disc}$ & 0.038 & 0.035 & 0.045 & 0.774 \\
             & Both$_{Mask}$  & 0.038 & 0.035 & 0.045 & 0.748 \\
\addlinespace
Inf$_{\Peers{}}$ & Both$_{Disc}$  & 0.037 & 0.034 & 0.038 & 0.855 \\
             & Bio$_{Disc}$   & 0.037 & 0.034 & 0.038 & 0.842 \\
             & Turns$_{Disc}$ & 0.037 & 0.034 & 0.038 & 0.825 \\
             & Both$_{Mask}$  & 0.037 & 0.034 & 0.038 & 0.755 \\
\bottomrule
\end{tabular}
\caption{NoMaskLoss trained without interlocutor-biography visibility (Trn$_{\Strangers{}}$).}%
\label{tab:overall-nml-mii}
\end{table*}

\begin{table*}[!htbp]
\centering
\small
\begin{tabular}{@{}l rrrr@{}}
\toprule
& \textbf{Inf$_{\Strangers{}}$} & \textbf{Inf$_{\Fan}$} & \textbf{Inf$_{I+I-}$} & \textbf{Inf$_{\Peers{}}$} \\
\midrule
& \multicolumn{4}{c}{\textit{Accuracy (mean over eval. conditions)}} \\
\cmidrule(lr){2-5}
Familiar   & 0.864 & 0.879 & 0.864 & 0.875 \\
Unfamiliar & 0.759 & 0.794 & 0.732 & 0.777 \\
\midrule
& \multicolumn{4}{c}{\textit{METEOR — Familiar}} \\
\cmidrule(lr){2-5}
Bio$_{Trg}$$\to$Turn$_{Trg}$ & 0.038 & 0.037 & 0.037 & 0.036 \\
Bio$_{Int}$$\to$Turn$_{Trg}$ & 0.044 & 0.043 & 0.036 & 0.036 \\
Bio$_{Trg}$$\to$Turn$_{Int}$ & 0.046 & 0.038 & 0.046 & 0.038 \\
\midrule
& \multicolumn{4}{c}{\textit{METEOR — Unfamiliar}} \\
\cmidrule(lr){2-5}
Bio$_{Trg}$$\to$Turn$_{Trg}$ & 0.037 & 0.036 & 0.038 & 0.037 \\
Bio$_{Int}$$\to$Turn$_{Trg}$ & 0.040 & 0.040 & 0.035 & 0.034 \\
Bio$_{Trg}$$\to$Turn$_{Int}$ & 0.045 & 0.038 & 0.045 & 0.038 \\
\bottomrule
\end{tabular}
\caption{Speaker familiarity for NoMaskLoss / Trn$_{\Strangers{}}$.}%
\label{tab:fam-nml-mii}

\end{table*}

\begin{table*}[!htbp]
\centering
\small
\begin{tabular}{@{}l rrrr@{}}
\toprule
& \textbf{Inf$_{\Strangers{}}$} & \textbf{Inf$_{\Fan}$} & \textbf{Inf$_{I+I-}$} & \textbf{Inf$_{\Peers{}}$} \\
\midrule
& \multicolumn{4}{c}{\textit{Accuracy (mean over eval. conditions)}} \\
\cmidrule(lr){2-5}
Familiar   & 0.820 & 0.847 & 0.804 & 0.842 \\
Unfamiliar & 0.792 & 0.819 & 0.777 & 0.802 \\
\bottomrule
\end{tabular}
\caption{Topic familiarity for NoMaskLoss / Trn$_{\Strangers{}}$.}%
\label{tab:topic-nml-mii}
\end{table*}

Tables~\ref{tab:overall-nml-mii}--\ref{tab:topic-nml-mii} show that removing the final-turn mask has only a modest effect when the model is trained in the \Strangers{} regime. Accuracy is very close to the MaskLoss Trn$_{\Strangers{}}$ counterpart, and the same inference configurations remain strongest. The overlap values are also similar, suggesting that in this training regime the dominant factor is not the loss objective alone but the absence of interlocutor-biography visibility during training. Speaker familiarity again explains more variance than topic familiarity, with unfamiliar pairings accounting for most of the accuracy loss.

\subsection{NoMaskLoss: Trn\texorpdfstring{$_{\Peers{}}$}{Peers}}

\begin{table*}[!htbp]
\centering
\small
\begin{tabular}{@{}llrrr r@{}}
\toprule
\textbf{Inf.} & \textbf{Eval} & \textbf{Bio$_{Trg}$$\to$Turn$_{Trg}$} & \textbf{Bio$_{Int}$$\to$Turn$_{Trg}$} & \textbf{Bio$_{Trg}$$\to$Turn$_{Int}$} & \textbf{Acc} \\
\midrule
Inf$_{\Strangers{}}$ & Both$_{Disc}$  & 0.032 & 0.036 & 0.038 & 0.824 \\
             & Bio$_{Disc}$   & 0.032 & 0.036 & 0.038 & 0.825 \\
             & Turns$_{Disc}$ & 0.032 & 0.036 & 0.038 & 0.757 \\
             & Both$_{Mask}$  & 0.032 & 0.036 & 0.038 & 0.707 \\
\addlinespace
Inf$_{\Fan}$ & Both$_{Disc}$  & 0.032 & 0.036 & 0.031 & 0.862 \\
             & Bio$_{Disc}$   & 0.032 & 0.036 & 0.031 & 0.830 \\
             & Turns$_{Disc}$ & 0.032 & 0.036 & 0.031 & 0.804 \\
             & Both$_{Mask}$  & 0.032 & 0.036 & 0.031 & 0.734 \\
\addlinespace
Inf$_{I+I-}$ & Both$_{Disc}$  & 0.032 & 0.030 & 0.037 & 0.809 \\
             & Bio$_{Disc}$   & 0.032 & 0.030 & 0.037 & 0.806 \\
             & Turns$_{Disc}$ & 0.032 & 0.030 & 0.037 & 0.736 \\
             & Both$_{Mask}$  & 0.032 & 0.030 & 0.037 & 0.710 \\
\addlinespace
Inf$_{\Peers{}}$ & Both$_{Disc}$  & 0.031 & 0.029 & 0.030 & 0.840 \\
             & Bio$_{Disc}$   & 0.031 & 0.029 & 0.030 & 0.819 \\
             & Turns$_{Disc}$ & 0.031 & 0.029 & 0.030 & 0.782 \\
             & Both$_{Mask}$  & 0.031 & 0.029 & 0.030 & 0.712 \\
\bottomrule
\end{tabular}
\caption{NoMaskLoss trained with mutual biography disclosure (Trn$_{\Peers{}}$).}%
\label{tab:overall-nml-miii}
\end{table*}

\begin{table*}[!htbp]
\centering
\small
\begin{tabular}{@{}l rrrr@{}}
\toprule
& \textbf{Inf$_{\Strangers{}}$} & \textbf{Inf$_{\Fan}$} & \textbf{Inf$_{I+I-}$} & \textbf{Inf$_{\Peers{}}$} \\
\midrule
& \multicolumn{4}{c}{\textit{Accuracy (mean over eval. conditions)}} \\
\cmidrule(lr){2-5}
Familiar   & 0.839 & 0.856 & 0.845 & 0.852 \\
Unfamiliar & 0.732 & 0.771 & 0.705 & 0.740 \\
\midrule
& \multicolumn{4}{c}{\textit{METEOR — Familiar}} \\
\cmidrule(lr){2-5}
Bio$_{Trg}$$\to$Turn$_{Trg}$ & 0.032 & 0.032 & 0.031 & 0.031 \\
Bio$_{Int}$$\to$Turn$_{Trg}$ & 0.038 & 0.038 & 0.031 & 0.031 \\
Bio$_{Trg}$$\to$Turn$_{Int}$ & 0.039 & 0.031 & 0.038 & 0.031 \\
\midrule
& \multicolumn{4}{c}{\textit{METEOR — Unfamiliar}} \\
\cmidrule(lr){2-5}
Bio$_{Trg}$$\to$Turn$_{Trg}$ & 0.032 & 0.031 & 0.032 & 0.031 \\
Bio$_{Int}$$\to$Turn$_{Trg}$ & 0.035 & 0.035 & 0.029 & 0.029 \\
Bio$_{Trg}$$\to$Turn$_{Int}$ & 0.037 & 0.030 & 0.037 & 0.030 \\
\bottomrule
\end{tabular}
\caption{Speaker familiarity for NoMaskLoss / Trn$_{\Peers{}}$.}%
\label{tab:fam-nml-miii}
\end{table*}

\begin{table*}[!htbp]
\centering
\small
\begin{tabular}{@{}l rrrr@{}}
\toprule
& \textbf{Inf$_{\Strangers{}}$} & \textbf{Inf$_{\Fan}$} & \textbf{Inf$_{I+I-}$} & \textbf{Inf$_{\Peers{}}$} \\
\midrule
& \multicolumn{4}{c}{\textit{Accuracy (mean over eval. conditions)}} \\
\cmidrule(lr){2-5}
Familiar   & 0.806 & 0.839 & 0.789 & 0.815 \\
Unfamiliar & 0.758 & 0.784 & 0.748 & 0.768 \\
\bottomrule
\end{tabular}
\caption{Topic familiarity for NoMaskLoss / Trn$_{\Peers{}}$.}%
\label{tab:topic-nml-miii}
\end{table*}

Tables~\ref{tab:overall-nml-miii}--\ref{tab:topic-nml-miii} give the NoMaskLoss model trained with mutual disclosure. Relative to MaskLoss Trn$_{\Peers{}}$, NoMaskLoss recovers some accuracy in several evaluation conditions, including the fully masked evaluation, but it does so with slightly higher biography--turn overlap. This is the trade-off motivating the main-paper choice of MaskLoss: standard next-token prediction can make the target easier to identify, but part of that gain comes from making biography text itself more predictable. The speaker and topic breakdowns show the same structure as before, with the largest gains under \Fan{} and \Peers{} inference and the largest residual difficulty on unfamiliar speakers.

\subsection{Zero-Shot Baseline}

\begin{table*}[!htbp]
\centering
\small
\begin{tabular}{@{}llrrr r@{}}
\toprule
\textbf{Inf.} & \textbf{Eval} & \textbf{Bio$_{Trg}$$\to$Turn$_{Trg}$} & \textbf{Bio$_{Int}$$\to$Turn$_{Trg}$} & \textbf{Bio$_{Trg}$$\to$Turn$_{Int}$} & \textbf{Acc} \\
\midrule
Inf$_{\Strangers{}}$ & Both$_{Disc}$  & 0.092 & 0.090 & 0.102 & 0.936 \\
             & Bio$_{Disc}$   & 0.092 & 0.090 & 0.102 & 0.963 \\
             & Turns$_{Disc}$ & 0.092 & 0.090 & 0.102 & 0.939 \\
             & Both$_{Mask}$  & 0.092 & 0.090 & 0.102 & 0.971 \\
\addlinespace
Inf$_{\Fan}$ & Both$_{Disc}$  & 0.093 & 0.091 & 0.095 & 0.952 \\
             & Bio$_{Disc}$   & 0.093 & 0.091 & 0.095 & 0.965 \\
             & Turns$_{Disc}$ & 0.093 & 0.091 & 0.095 & 0.954 \\
             & Both$_{Mask}$  & 0.093 & 0.091 & 0.095 & 0.971 \\
\addlinespace
Inf$_{I+I-}$ & Both$_{Disc}$  & 0.089 & 0.083 & 0.100 & 0.930 \\
             & Bio$_{Disc}$   & 0.089 & 0.083 & 0.100 & 0.954 \\
             & Turns$_{Disc}$ & 0.089 & 0.083 & 0.100 & 0.927 \\
             & Both$_{Mask}$  & 0.089 & 0.083 & 0.100 & 0.955 \\
\addlinespace
Inf$_{\Peers{}}$ & Both$_{Disc}$  & 0.089 & 0.083 & 0.093 & 0.953 \\
             & Bio$_{Disc}$   & 0.089 & 0.083 & 0.093 & 0.957 \\
             & Turns$_{Disc}$ & 0.089 & 0.083 & 0.093 & 0.947 \\
             & Both$_{Mask}$  & 0.089 & 0.083 & 0.093 & 0.962 \\
\bottomrule
\end{tabular}
\caption{Zero-shot baseline (M$_{z\text{-}s}$).}%
\label{tab:overall-zeroshot}
\end{table*}

\begin{table*}[!htbp]
\centering
\small
\begin{tabular}{@{}l rrrr@{}}
\toprule
& \textbf{Inf$_{\Strangers{}}$} & \textbf{Inf$_{\Fan}$} & \textbf{Inf$_{I+I-}$} & \textbf{Inf$_{\Peers{}}$} \\
\midrule
& \multicolumn{4}{c}{\textit{Accuracy (mean over eval. conditions)}} \\
\cmidrule(lr){2-5}
Familiar   & 0.993 & 0.996 & 0.996 & 0.996 \\
Unfamiliar & 0.894 & 0.919 & 0.880 & 0.921 \\
\midrule
& \multicolumn{4}{c}{\textit{METEOR — Familiar}} \\
\cmidrule(lr){2-5}
Bio$_{Trg}$$\to$Turn$_{Trg}$ & 0.092 & 0.092 & 0.085 & 0.084 \\
Bio$_{Int}$$\to$Turn$_{Trg}$ & 0.093 & 0.094 & 0.083 & 0.083 \\
Bio$_{Trg}$$\to$Turn$_{Int}$ & 0.100 & 0.091 & 0.096 & 0.088 \\
\midrule
& \multicolumn{4}{c}{\textit{METEOR — Unfamiliar}} \\
\cmidrule(lr){2-5}
Bio$_{Trg}$$\to$Turn$_{Trg}$ & 0.093 & 0.093 & 0.091 & 0.092 \\
Bio$_{Int}$$\to$Turn$_{Trg}$ & 0.089 & 0.089 & 0.083 & 0.083 \\
Bio$_{Trg}$$\to$Turn$_{Int}$ & 0.103 & 0.097 & 0.101 & 0.096 \\
\bottomrule
\end{tabular}
\caption{Speaker familiarity for the zero-shot baseline.}%
\label{tab:fam-zeroshot}
\end{table*}

\begin{table*}[!htbp]
\centering
\small
\begin{tabular}{@{}l rrrr@{}}
\toprule
& \textbf{Inf$_{\Strangers{}}$} & \textbf{Inf$_{\Fan}$} & \textbf{Inf$_{I+I-}$} & \textbf{Inf$_{\Peers{}}$} \\
\midrule
& \multicolumn{4}{c}{\textit{Accuracy (mean over eval. conditions)}} \\
\cmidrule(lr){2-5}
Familiar   & 0.955 & 0.964 & 0.938 & 0.953 \\
Unfamiliar & 0.950 & 0.958 & 0.944 & 0.956 \\
\bottomrule
\end{tabular}
\caption{Topic familiarity for the zero-shot baseline.}%
\label{tab:topic-zeroshot}
\end{table*}

Tables~\ref{tab:overall-zeroshot}--\ref{tab:topic-zeroshot} show why the zero-shot baseline should be interpreted differently from the fine-tuned models. Its accuracy is near ceiling across evaluation conditions, but its METEOR overlap values are much higher than those of the gold dialogues and the fine-tuned systems. This means that high recognisability is accompanied by strong surface similarity to the biographies. Speaker familiarity still matters, especially for unfamiliar pairings, but the topic-familiarity split is almost flat: once the zero-shot model copies enough biographical material, the topic itself contributes little additional diagnostic structure.

\section{Rare-Word Leakage Subsets}
\label{app:RQ3-subsets}

Section~\ref{sec:RQ3-results} reports the within configuration accuracy gain when interlocutor turns contain at least one rare word from the target biography (Figure~\ref{fig:RQ3-uplift}). Table~\ref{tab:RQ3-uplift-subsets} reports the absolute subset accuracies underlying that gain, together with subset sizes.

Overlap-positive subsets are near ceiling in every configuration (0.968--1.000), confirming that exact rare-word overlap is a strong identification cue. Under Trn$_\Strangers{}$, however, the no-overlap subset is also more accurate under \Fan{} than under \Strangers{} (Turns$_\text{Disc}$: 0.839 vs.\ 0.785; Both$_\text{Disc}$: 0.869 vs.\ 0.827), suggesting that rare-word overlap is a lower bound on broader leakage through paraphrase, topic steering, or implicit semantic uptake.

\begin{table}[t]
\centering
\small
\setlength{\tabcolsep}{4pt}
\begin{tabular}{@{}lrrr@{}}
\toprule
\textbf{Eval} & \textbf{No ov.} & \textbf{Ov.} & \textbf{$\Delta$ pp} \\
\midrule
\multicolumn{4}{@{}l}{\textit{M$_{z-s}$ / Inf$_\Strangers{}$}} \\
\quad Both$_\text{Disc}$  & 0.919 (3123) & 0.980 (1252) & $+6.1$ \\
\quad Turns$_\text{Disc}$ & 0.921 (3123) & 0.984 (1252) & $+6.3$ \\
\addlinespace
\multicolumn{4}{@{}l}{\textit{M$_{z-s}$ / Inf$_\Fan{}$}} \\
\quad Both$_\text{Disc}$  & 0.935 (2992) & 0.988 (1383) & $+5.2$ \\
\quad Turns$_\text{Disc}$ & 0.937 (2992) & 0.991 (1383) & $+5.4$ \\
\midrule
\multicolumn{4}{@{}l}{\textit{Trn$_\Strangers{}$ / Inf$_\Strangers{}$}} \\
\quad Both$_\text{Disc}$  & 0.827 (4280) & 0.968 (95) & $+14.1$ \\
\quad Turns$_\text{Disc}$ & 0.785 (4280) & 0.979 (95) & $+19.4$ \\
\addlinespace
\multicolumn{4}{@{}l}{\textit{Trn$_\Strangers{}$ / Inf$_\Fan{}$}} \\
\quad Both$_\text{Disc}$  & 0.869 (4240) & 1.000 (135) & $+13.1$ \\
\quad Turns$_\text{Disc}$ & 0.839 (4240) & 0.985 (135) & $+14.6$ \\
\bottomrule
\end{tabular}
\caption{Judge accuracy split by whether interlocutor turns contain any exact
rare-word overlap with the target biography. Parentheses give subset sizes;
$\Delta$ is overlap minus no overlap accuracy in percentage points.}
\label{tab:RQ3-uplift-subsets}
\end{table}

\section{Training--Inference Dialogue Examples}
\label{app:dialogue-examples}

All examples use Walter Sobchak from \textit{The Big Lebowski} (Speaker~1, target) and Peter Parker / Spider-Man from \textit{Spider-Man} (Speaker~2, interlocutor), enabling direct comparison across training--inference conditions. These examples are illustrative; the claims in the main paper rest on aggregate results. Both biographies, reproduced verbatim from PRODIGy~\citep{occhipinti-etal-2024-prodigy}, are reported in Table~\ref{tab:app-exp-RQ3-biographies}; the four configuration tables that follow refer back to them.

\begin{center}
\begin{minipage}{\columnwidth}
\begin{mdframed}[backgroundcolor=dialoguebg, linecolor=dialogueframe, linewidth=1.2pt, roundcorner=5pt, innerleftmargin=10pt, innerrightmargin=10pt, innertopmargin=8pt, innerbottommargin=8pt, nobreak=true]
\footnotesize\raggedright
\textbf{Target biography (Walter):} I am owner of Sobchak Security; I use the combat knowledge I accumulated during the Vietnam War. I spend most of my time at the local bowling alley, and picking fights with those who don't respect the rules. I am stubborn, aggressive, and a mentally unhinged believer in rules. I have to help my friend The Dude through a sticky situation. While my tour of duty prepared me for a career in security, it also left me paranoid and angry.\\[0.6em]
\textbf{Interlocutor biography (Peter / Spider-Man):} I am a superhero and \underline{freelance photographer} taking and selling \underline{pictures} of myself as \copied{Spider-Man}. I am kind, reserved, clever, a lifelong outcast and I tend to be reserved and quiet. Whenever I am \copied{Spider-Man}, I become a consummate jokester using one-liners to irritate my foes. I am super strong, super agile, capable of sensing danger, and able to shoot \copied{webs} out of my hands. I have to keep New York City safe.
\end{mdframed}
\captionsetup{hypcap=false}
\captionof{table}{PRODIGy biographies for the target (Walter) and interlocutor (Peter / Spider-Man) used throughout this appendix, shortened to the five most representative sentences each. Highlighted spans anticipate the dialogues below: \copied{lexical copying} (Table~\ref{tab:app-exp-RQ3-concordant-minus}) and \underline{thematic adaptation} (Table~\ref{tab:app-exp-RQ3-concordant-plus}).}
\label{tab:app-exp-RQ3-biographies}
\end{minipage}
\end{center}

Tables~\ref{tab:app-exp-RQ3-concordant-minus}--\ref{tab:app-exp-RQ3-concordant-plus} illustrate all four training$\times$inference configurations. Concordant Trn$_{\Strangers{}}$/Inf$_{\Strangers{}}$ shows Spider-Man-related terms surfacing despite masking. Concordant Trn$_{\Peers{}}$/Inf$_{\Peers{}}$ shows contextual adaptation to the interlocutor. Discordant Trn$_{\Strangers{}}$/Inf$_{\Peers{}}$ does not visibly use the newly supplied interlocutor information, while discordant Trn$_{\Peers{}}$/Inf$_{\Strangers{}}$ produces generic output when the expected interlocutor context is absent.

\begin{center}
\begin{minipage}{\columnwidth}
\begin{mdframed}[backgroundcolor=dialoguebg, linecolor=dialogueframe, linewidth=1.2pt, roundcorner=5pt, innerleftmargin=10pt, innerrightmargin=10pt, innertopmargin=8pt, innerbottommargin=8pt, nobreak=true]
\textbf{Trn$_{\Strangers{}}$/Inf$_{\Strangers{}}$ [Concordant: both WITHOUT]} \\[0.4em]
\footnotesize\raggedright
\textbf{Target Bio (Walter):} as in Table~\ref{tab:app-exp-RQ3-biographies}.\\[0.2em]
\textbf{Interlocutor Bio (Peter):} [MASKED] (contains \copied{Spider-Man}, \copied{webs}) \textbf{Topic:} Fiction\\[0.2em]\hrule\vspace{0.3em}
\spkone{What? You mean like Jim?}
\spktwo{\copied{Spiderman}!}
\spkone{Yeah\ldots Like \copied{Spiderman}!}
\spktwo{How do you know it's me?!}
\spkone{Because you've got these\ldots little \copied{web} things on your clothes! Yeah -- there's one right here\ldots}
\end{mdframed}
\captionsetup{hypcap=false}
\captionof{table}{Concordant no-disclosure setting (Trn$_{\Strangers{}}$/Inf$_{\Strangers{}}$). Spider-Man-related terms still surface in both speakers' turns (\copied{highlighted}), illustrating residual lexical leakage.}
\label{tab:app-exp-RQ3-concordant-minus}
\end{minipage}
\end{center}

\begin{center}
\begin{minipage}{\columnwidth}
\begin{mdframed}[backgroundcolor=dialoguebg, linecolor=dialogueframe, linewidth=1.2pt, roundcorner=5pt, innerleftmargin=10pt, innerrightmargin=10pt, innertopmargin=8pt, innerbottommargin=8pt, nobreak=true]
\textbf{Trn$_{\Strangers{}}$/Inf$_{\Peers{}}$ [Discordant: train WITHOUT, infer WITH]} \\[0.4em]
\footnotesize\raggedright
\textbf{Target Bio (Walter):} as in Table~\ref{tab:app-exp-RQ3-biographies}.\\[0.2em]
\textbf{Interlocutor Bio (Peter):} as in Table~\ref{tab:app-exp-RQ3-biographies}. \textbf{Topic:} Fiction\\[0.2em]\hrule\vspace{0.3em}
\spkone{They didn't tell me it was a limited engagement! Look here -- he's getting ready for another one now\ldots}
\spktwo{Shut up about that stuff right now. Let's talk turkey here. What were you doing there?}
\spkone{Did you see anything weird tonight?}
\spktwo{Hey, what're we talking about? No way I'm going back over all this again\ldots}
\end{mdframed}
\captionsetup{hypcap=false}
\captionof{table}{Discordant disclosure setting (Trn$_{\Strangers{}}$/Inf$_{\Peers{}}$). The model receives the interlocutor biography at inference, but the supplied Spider-Man context is not visibly integrated.}
\label{tab:app-exp-RQ3-discordant-minus-plus}
\end{minipage}
\end{center}

\begin{center}
\begin{minipage}{\columnwidth}
\begin{mdframed}[backgroundcolor=dialoguebg, linecolor=dialogueframe, linewidth=1.2pt, roundcorner=5pt, innerleftmargin=10pt, innerrightmargin=10pt, innertopmargin=8pt, innerbottommargin=8pt, nobreak=true]
\textbf{Trn$_{\Peers{}}$/Inf$_{\Strangers{}}$ [Discordant: train WITH, infer WITHOUT]} \\[0.4em]
\footnotesize\raggedright
\textbf{Target Bio (Walter):} as in Table~\ref{tab:app-exp-RQ3-biographies}.\\[0.2em]
\textbf{Interlocutor Bio (Peter):} [MASKED] \textbf{Topic:} Fiction\\[0.2em]\hrule\vspace{0.3em}
\spkone{Listen very carefully -- I'm going through this once more for you. This man has become like a son to me\ldots}
\spktwo{That name sounds familiar somehow.}
\spkone{It's probably because he was an engineer on one of our planes shot down over Laos\ldots}
\spktwo{What's wrong?}
\end{mdframed}
\captionsetup{hypcap=false}
\captionof{table}{Discordant masking setting (Trn$_{\Peers{}}$/Inf$_{\Strangers{}}$). A model trained to expect interlocutor context produces generic dialogue when that context is absent at inference.}
\label{tab:app-exp-RQ3-discordant-plus-minus}
\end{minipage}
\end{center}

\begin{center}
\begin{minipage}{\columnwidth}
\begin{mdframed}[backgroundcolor=dialoguebg, linecolor=dialogueframe, linewidth=1.2pt, roundcorner=5pt, innerleftmargin=10pt, innerrightmargin=10pt, innertopmargin=8pt, innerbottommargin=8pt, nobreak=true]
\textbf{Trn$_{\Peers{}}$/Inf$_{\Peers{}}$ [Concordant: both WITH]} \\[0.4em]
\footnotesize\raggedright
\textbf{Target Bio (Walter):} as in Table~\ref{tab:app-exp-RQ3-biographies}.\\[0.2em]
\textbf{Interlocutor Bio (Peter):} as in Table~\ref{tab:app-exp-RQ3-biographies}. \textbf{Topic:} Fiction\\[0.2em]\hrule\vspace{0.3em}
\spkone{Oh boy! Did you see? He's gonna get it again\ldots What the hell was wrong with him?}
\spktwo{What does this thing do?}
\spkone{It takes a \underline{picture} when they hit me or something. Come check it out\ldots}
\spktwo{Can it keep up? My \underline{camera} could make money doing this all by itself!}
\end{mdframed}
\captionsetup{hypcap=false}
\captionof{table}{Concordant mutual-disclosure setting (Trn$_{\Peers{}}$/Inf$_{\Peers{}}$). Aligned training and inference produce thematic adaptation to Spider-Man's photographer context (underlined).}
\label{tab:app-exp-RQ3-concordant-plus}
\end{minipage}
\end{center}

\end{document}